\pdfoutput=1

\documentclass[11pt]{article}

\usepackage[]{acl}

\usepackage{times}
\usepackage{latexsym}

\usepackage[T1]{fontenc}

\usepackage[utf8]{inputenc}

\usepackage{microtype}

\usepackage{algorithm}
\usepackage{inconsolata}
\usepackage{mathrsfs}
\usepackage{amssymb}
\usepackage{amsmath}
\usepackage{graphicx}
\usepackage{colortbl}
\usepackage{booktabs}
\usepackage{multirow}
\usepackage{arydshln}
\usepackage{textpos}
\usepackage{algpseudocode}
\usepackage{subfigure} 
\usepackage{textpos}

\DeclareMathOperator*{\argmax}{argmax}
\algnewcommand\algorithmicinput{\textbf{Input:}}
\algnewcommand\INPUT{\item[\algorithmicinput]}
\algnewcommand\algorithmicoutput{\textbf{Output:}}
\algnewcommand\OUTPUT{\item[\algorithmicoutput]}
\algnewcommand\algorithmicbegin{\textbf{begin}}
\algnewcommand\BEGIN{\item[\algorithmicbegin]}
\algnewcommand\algorithmicendbegin{\textbf{end}}
\algnewcommand\ENDBEGIN{\item[\algorithmicendbegin]}

\algdef{SE}[DOWHILE]{Do}{doWhile}{\algorithmicdo}[1]{\algorithmicwhile\ #1}%

\newcommand{\TheName}{FibecFed}

\title{Fisher Information-based Efficient Curriculum Federated Learning with Large Language Models}

\author{
        \setcounter{footnote}{1}
        Ji Liu$^{1}$\thanks{\ \ ~~~Equal contribution. }$^{\ast}$,
        \setcounter{footnote}{0}
        Jiaxiang Ren$^{2\dag}$,
        Ruoming Jin$^{3}$,
        Zijie Zhang$^4$, \\
        \textbf{Yang Zhou$^{2}$\thanks{\ Corresponding author: jiliuwork@gmail.com, yangzhou@auburn.edu},
        Patrick Valduriez$^{5}$,
        Dejing Dou$^{6}$}\\
        \small $^1$HiThink Research, Hangzhou, Zhejiang, China,\\
        \small $^2$Auburn University, Auburn, United States,
        \small $^3$Kent State University, Kent, United States,\\
        \small $^4$University of Texas at San Antonio, San Antonio, United States,\\
        \small $^5$Inria, University of Montpellier, CNRS, LIRMM, France, and LNCC, Petropolis, Brazil,\\
        \small $^6$Fudan University, Shanghai, China, and BEDI Cloud, Beijing, China \\
}

\newcommand{\rev}[1]{\textcolor{black}{#1}}

\begin{document}
\maketitle
\begin{abstract}
As a promising paradigm to collaboratively train models with decentralized data, Federated Learning (FL) can be exploited to fine-tune Large Language Models (LLMs).
While LLMs correspond to huge size, the scale of the training data significantly increases, which leads to tremendous amounts of computation and communication costs. The training data is generally non-Independent and Identically Distributed (non-IID), which requires adaptive data processing within each device. Although Low-Rank Adaptation (LoRA) can significantly reduce the scale of parameters to update in the fine-tuning process, it still takes unaffordable time to transfer the low-rank parameters of all the layers in LLMs. In this paper, we propose a Fisher Information-based Efficient Curriculum Federated Learning framework (\TheName{}) with two novel methods, i.e., adaptive federated curriculum learning and efficient sparse parameter update. First, we propose a fisher information-based method to adaptively sample data within each device to improve the effectiveness of the FL fine-tuning process. Second, we dynamically select the proper layers for global aggregation and sparse parameters for local update with LoRA so as to improve the efficiency of the FL fine-tuning process. Extensive experimental results based on 10 datasets demonstrate that \TheName{} yields excellent performance (up to 45.35\% in terms of accuracy) and superb fine-tuning speed (up to 98.61\% faster) compared with 17 baseline approaches). 
\end{abstract}

\section{Introduction}

As a promising paradigm to collaboratively train models with decentralized data, Federated Learning (FL) can be exploited to fine-tune Large Language Models (LLMs) without aggregating the raw data from a large number of devices \cite{fan2023fatellm,kuang2023federatedscopellm,che-etal-2023-federated,liu2024enhancing,che2023fast,liu2022multi,liu2022distributed,zhou2022efficient}. A number of stringent legal regulations \cite{GDPR,CCPA} have been set up in order to protect the security and the privacy of personal data, which hinders the aggregation of the decentralized raw data. FL typically utilizes a parameter server \cite{li2014scaling,liu2024fedasmu,liu2023heterps,liu2023distributed} to aggregate the distributed model updates in devices, which only transfers the parameters or the gradients of the updated models in replace of the raw personal data. \rev{By leveraging the distributed raw data of end users, Large Language Models (LLMs) can be trained on devices with excellent performance \cite{zhao2024llmbased}.}

\begin{textblock*}{8cm}(-3.5cm,12.5cm) 
    {\LARGE To appear in EMNLP 2024}
\end{textblock*}

While ChatGPT \cite{ChatGPT} has achieved remarkable progress, LLMs \cite{touvron2023llama,jiang2023mistral,du2022glm,zeng2023glm-130b,zhang2024tinyllama} have attracted extensive attention. The size of LLMs ranges from several million parameters, e.g., RoBERTa\textsubscript{LARGE} \cite{Liu2020RoBERTa}, to several hundreds billion parameters \cite{wang2019language}. While the large scale brings strong capability in various Natural Language Processing (NLP) tasks \cite{zhao2023survey}, the pre-training and the fine-tuning process of LLMs  significant communication and computation costs \cite{geminiteam2023gemini}. 

Two types of parameter-efficient approaches exist for reducing the number of parameters within the fine-tuning process of LLMs. Prompt tuning \cite{liu2021gpt,Lester2021Power,liu2022p} can dynamically adjust the prompts to fine-tune LLMs with only a few trainable parameters, which may introduce performance degradation. Although Low-Rank Adaptation (LoRA) \cite{hu2021lora} can significantly reduce the scale of parameters to update in the fine-tuning process of LLMs \rev{so as to enable the training of LLMs on edge devices \cite{xu2024fwdllm}}, it still takes unaffordable time to update the low-rank parameters of all the layers in LLMs when dealing with decentralized data. 

While being an effective method to improve the efficiency and effectiveness of training process, curriculum learning \cite{Bengio2009CL} is exploited to train large-scale models \cite{li2022the}. Inspired by the learning strategy of starting small \cite{ELMAN199371}, the curriculum training process starts with easier data and then gradually increase the difficulty. Instead of randomly sampling the batch from training dataset, curriculum learning allows the model to gradually learn from easy samples to hard samples during the training or the fine-tuning process. Existing approaches generally measure the complexity of samples based on heuristic methods \cite{li2024deepspeed} or a simple mode-based method \cite{xu2022modelbased}, both of which cannot provide an accurate estimation of difficulty of data samples and cannot be directly applied in FL. In the context of FL, the data is generally non-Independent and Identically Distributed (non-IID), which requires an adaptive difficulty evaluation approach for diverse devices \cite{Vahidian2023When}. 

Model compression methods, e.g., pruning \cite{Wu2021FedSCR,liu2024efficient,Zhang2022FedDUAP} or sparse training \cite{Bibikar2022Federated}, are exploited in FL to reduce computation and communication costs. Sparse training can achieve personalization so as to further improve the performance of FL \cite{liu2023sparse,Dai2022DisPFLTC}. However, the pruning or sparse training incurs severe accuracy degradation due to lossy strategies or simple component (neuron) selection mechanisms. In addition, the model can be split into two parts, i.e., the server part and the device part, in order to achieve both the generalization and personalization capability \cite{han2023splitgp}. 

Fisher information can be exploited to accelerate the training process of LLMs \cite{ollivier2015riemannian,martens2015optimizing,osawa2023pipefisher}. Fisher information is defined as the amount of information carried by a random variable corresponding to some unknown parameters \cite{duy2022fisher}. As a measure of the local curvature \cite{martens2020new}, Fisher Information Matrix (FIM) defines the Riemannian metric of the parameter space \cite{karakida2019universal}, which can indicate the difficulty of data samples and the importance of each component of the network along with the generalization performance \cite{jastrzebski2021catastrophic}. 

In this paper, we propose \TheName{}, i.e., a Fisher Information-based Efficient Curriculum Federated Learning framework. \TheName{} is composed of two novel methods, i.e., adaptive federated curriculum learning and efficient sparse parameter update. To the best of our knowledge, we are among the first to exploit the fisher information to perform curriculum learning and sparse training at the same time within FL settings. We summarize out major contributions as follows:

\begin{itemize}
    \item We propose an adaptive federated curriculum learning method to sample easy data samples first and to gradually improve the difficulty of samples so as to improve the effectiveness of the FL fine-tuning process. We exploit a fisher information-based method to measure the difficulty of training data within each device.  
    
    \item We propose an efficient sparse parameter update method to select proper layers for global aggregation and to adaptively update sparse parameters to achieve excellent efficiency and effectiveness. We utilize fisher information to evaluate the importance of diverse components of LLMs and propose a lossless method for global aggregation and local update.

    \item We conduct extensive experimentation to validate our approach using 10 datasets. The experimental results reveal that \TheName{} significantly outperforms 17 baseline approaches in terms of accuracy (up to 45.35\% higher) and fine-tuning speed (up to 98.61\% faster). 

\end{itemize}

The rest of the paper is organized as follows. The related work is presented in Section \ref{sec:relatedWork}. We formulate the problem to address in Section \ref{sec:problem}. We present the architecture of \TheName{} and propose the adaptive federated curriculum learning and the efficient sparse parameter update method in Section \ref{sec:effCuFL}. We demonstrate the experimental results in Section \ref{sec:exp}. Finally, Section \ref{sec:con} concludes.

\section{Related Word \& Preliminaries}
\label{sec:relatedWork}

Inspired by the learning strategy of starting small \cite{ELMAN199371}, curriculum learning \cite{Bengio2009CL} is exploited in large-scale model training \cite{li2022the}. Existing works measure the complexity of samples based on static characteristics of data samples, e.g., sequence length  \cite{li2024deepspeed,platanios2019Competence}. Although a simple global mode-based method is proposed to predict the performance improvement based on several training states \cite{xu2022modelbased}, it still cannot provide an accurate estimation of difficulty of data samples due to non-IID data in FL settings. Direct evaluation based on the inference loss of models \cite{Vahidian2023When} cannot well explore the impact on the generalization of the training process. The attention scores can analyze the dependency among diverse layers, but varies significantly between heads \cite{vig2019analyzing}, which cannot be directly utilized in FL settings. While a sharpness-aware minimization method \cite{foret2021sharpnessaware} can help minimize loss value and loss sharpness to improve model generalization, it does not consider federated fine-tuning settings of LLMs. 

\rev{FIM can be exploited to enable the second-order optimization so as to improve the training process \cite{osawa2023pipefisher,jin2022accelerated} and to compute a global posterior for federated learning \cite{jhunjhunwala2024fedfisher}. In addition, continual learning can be used to improve the performance of trained models while addressing the forgetting problem \cite{wu2022pretrained}. Different from \cite{osawa2023pipefisher,wu2022pretrained,jhunjhunwala2024fedfisher}, we exploit the sum of diagonal of FIM to evaluate the difficulty of samples within the efficient curriculum learning method and to calculate the importance score of each layer and neuron within the LLM.}

Model compression methods \cite{Wu2021FedSCR,Bibikar2022Federated} are exploited in FL to reduce both computation and communication costs. Although pruning methods can reduce the size of large models \cite{wang-etal-2020-structured, ma2023llmpruner,xia2023sheared}, it is complicated to choose a proper pruning rate and may incur inferior performance in terms of accuracy \cite{Wu2021FedSCR}.  Sparse training can achieve personalization \cite{Bibikar2022Federated,liu2023sparse,setayesh2022perfedmask,Dai2022DisPFLTC} while addressing the client shift problem brought by the non-IID data \cite{setayesh2022perfedmask,Karimireddy2020SCAFFOLD}. However, the existing sparse training methods may incur severe accuracy degradation with poor generalization capacity due to simple component selection mechanisms. The model can be split into a server part and a device part to achieve both generalization and personalization capability \cite{han2023splitgp}, which still incurs severe computation and communication costs in the FL settings of LLMs. \rev{Please note our approach is orthogonal with model compression methods.}

For NLP tasks, prompt tuning \cite{liu2021gpt,Lester2021Power,liu2022p} can fine-tune LLMs with only a few parameters. With prompt tuning, an extra network is exploited to generate proper prompts or prefix, which is concatenated with the input to guide LLMs to generate proper answers. Furthermore, LoRA updates trainable rank decomposition matrices while freezing the parameters of the original network, which can significantly reduce the scale of parameters to update \cite{hu2021lora}. However, both the prompt tuning and LoRA still incur significant communication costs due to the update for all the layers. 

\section{Problem Formulation}
\label{sec:problem}

In this paper, we delve into the problem of how to efficiently fine-tune a large language model within a FL setting. The FL setting is composed of a parameter server and $K$ devices. We assume that the data samples are distributed among the devices, each of which contains a dataset $D_k = \{s_i, m_i \}^{n_k}$ with $s_i$, $m_i$, and $n_k$ referring to a data sample, the corresponding label, and the cardinality of $D_k$. We denote the cardinality of the whole dataset $D = \{D_1, D_2, ..., D_k\}$ 
by $N$. 

We consider a LLM $\mathcal{M}$ of $L$ layers, each layer contains a full parameter matrix $\mathcal{W}_{o}^l$. We exploit the LoRA method to reduce the parameters to update in this paper \cite{hu2021lora}, and denote the LoRA parameters of the the LLM $\mathcal{M}$ by $P$ with $P_l$ representing the set of LoRA parameters in Layer $l$ of $\mathcal{M}$. 
We denote the updated LoRA parameters on Device $k$ by $\mathcal{P}_k$.
Then, we formulate the problem to address in this paper as how to efficiently update $\mathbf{P}$ so as to minimize the global loss:
\begin{equation}
\min_{\mathbf{P}}\left[\mathcal{F}(\mathcal{M}, \mathbf{P})\triangleq\frac{1}{K}\sum_{k = 1, ~\mathcal{P}_k \in \mathbf{P}}^K n_k F_k(\mathcal{M}, \mathcal{P}_k)\right],
\label{eq:problem}
\end{equation}
where $\mathcal{F}(\mathcal{M}, \mathbf{P})$ is the global loss, $F_i(\mathcal{M}, \mathcal{P}_k)\triangleq\frac{1}{n_l}\sum_{\{s_i, m_i\} \in \mathcal{D}_k} f(\mathcal{M}, \mathcal{P}_k, s_i, m_i)$ represents the local loss function on Device $k$ with $f(\mathcal{M}, \mathcal{P}_k, s_i, m_i)$ calculating the local loss on Device $k$.



\begin{figure*}[!t]
\centering
\includegraphics[width=\linewidth]{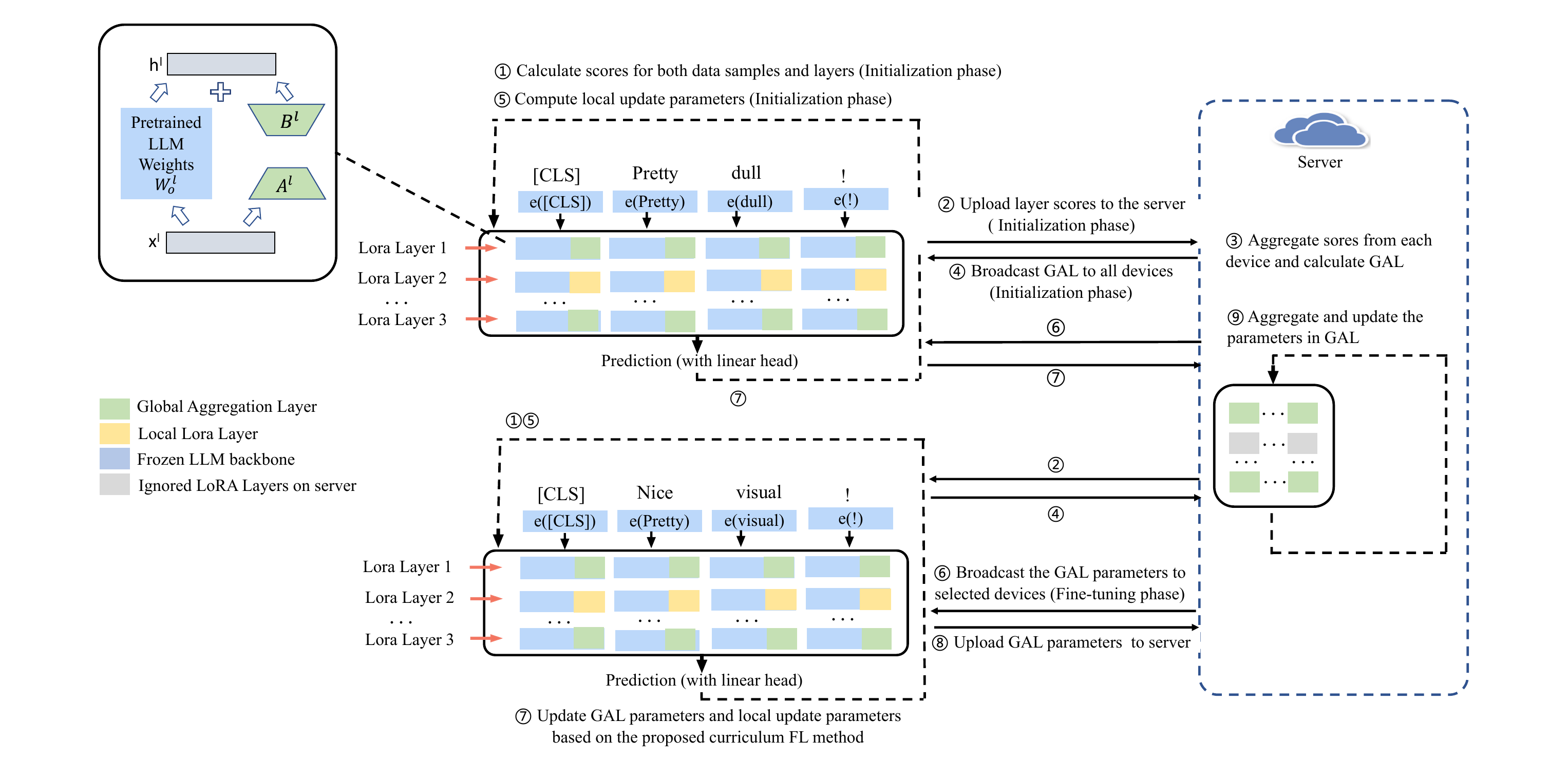}
\caption{The system model of \TheName{}.}
\label{fig:framework}
\end{figure*}

\section{Efficient Curriculum Federated Learning}
\label{sec:effCuFL}

In this section, we first explain the system model. Then, we propose the adaptive federated curriculum learning method. Afterward, we further detail the efficient sparse tuning method. 

\subsection{System Model}
\label{sec:SystemModel}

The system model of \TheName{} is shown in Figure \ref{fig:framework}. We assume that the LLM ($\mathcal{M}$) is deployed on each device. 
The parameters of $\mathcal{M}$ stays frozen while we update the LoRA parameters \cite{hu2021lora}. As shown on the top left of Figure \ref{fig:framework}, on Device $k$, the parameters ($W_o^l$) at each layer ($l$) is decomposed into two matrices (LoRA), i.e., $A^l_k$ and $B^l_k$, which can be updated during the fine-tuning process. Then, the hidden values ($h$) generated at Layer $l$ with the input $x$ is calculated based on Formula \ref{eq:lora}.
\begin{equation}
h = W_o^l x + B^l_k A^l_k x.
\label{eq:lora}
\end{equation}

The fine-tuning process is composed of two phases, i.e., initialization and tuning. Within the initialization phase, we evaluate the difficulty score for each batch of data samples (see Formulas \ref{eq:FIM}-\ref{eq:appFIM} in Section \ref{subsec:cl} and Lines \ref{line:initialBegin}-\ref{line:scoreEnd} in Algorithm \ref{alg:ficfl}) and the importance score for each layer (see Formulas \ref{eq:noise}-\ref{eq:layerScoreGlobal} and details in Section \ref{subsec:sparseLayerTraining}) based on fisher information on each device (Step \textcircled{1}). Then, the importance score of each layer is transferred to the server (Step \textcircled{2}), which aggregates the scores and selects proper layers as the Global Aggregation Layers (GAL) (see details in Section \ref{subsec:sparseLayerTraining}, Line \ref{line:gal} in Algorithm \ref{alg:ficfl}, Step \textcircled{3}). Afterward, the GAL are boradcasted to all the devices (Step \textcircled{4}). Finally, the parameters, which are not in the GAL, are locally evaluated on each device so as to generate local update part of parameters and the local static parameters to be frozen (see details in Section \ref{subsec:pupdate}, Lines \ref{line:pupdateFor}-\ref{line:initialEnd} in Algorithm \ref{alg:ficfl}, Step \textcircled{5}). During the tuning phase, only the parameters in GAL and the local update part of parameters are updated while the local static parameters are kept frozen. The tuning phase consists of multiple rounds, each of which consists of five steps. First, the server randomly selects $\mathcal{K}$ devices and broadcasts the global parameters in GAL on the server to the selected devices (Step \textcircled{6}, Line \ref{line:deviceSampling} in Algorithm \ref{alg:ficfl}). Then, the parameters in GAL and the local update part of parameters are updated based on our proposed curriculum FL (see details in Appendix) on each selected device (Step \textcircled{7}, Lines \ref{line:localUpdateBegin}-\ref{line:localUpdateEnd} in Algorithm \ref{alg:ficfl}). Afterward, the updated parameters in GAL are uploaded to the server (Step \textcircled{8}). Finally, the server aggregate and update the global parameters in GAL (Step \textcircled{9}, Line \ref{line:tuningEnd} in Algorithm \ref{alg:ficfl}).

\subsection{Fisher information-based Curriculum Federated Learning}
\label{subsec:cl}

Inspired by the starting small strategy \cite{ELMAN199371}, we propose a fisher information-based curriculum FL method to enable efficient federated fine-tuning. As the FIM can help indicate the amount of information carried by each data sample to generate the response \cite{ly2017tutorial}, we propose utilizing the FIM to measure the difficulty of data samples. The FIM is defined in Formula \ref{eq:FIM}:
\begin{equation}
\label{eq:FIM}
\mathbf{F}_{i}  \triangleq 
\mathbb{E}_{s_i} \left[
    \left(
    \nabla \log p_{k}(s_i)
    \right)
    \left(
    \nabla \log p_{k}(s_i)
    \right)^T
\right],
\end{equation}
where $\mathbf{F}_{i}$ represents the FIM corresponding to data sample $s_i$, $p_{k}(s_i)$ represents the probability density function of the inference with the LLM $\mathcal{M}$, the LoRA parameters $\mathcal{P}_k$, and data sample $s_i$, $\nabla \log p_{k}(s_i)$ denotes the first-order derivative of the LoRA parameters, calculated by the gradient of the loss respect to $\mathcal{P}_k$\rev{, $T$ refers to the transpose of a matrix}. Practically, the expected FIM can be approximated by empirical FIM \cite{kunstner2019limitations} as defined in Formula \ref{eq:eFIM}:
\begin{equation}
\label{eq:eFIM}
\mathbf{F}_{i} \approx \frac{1}{N}\sum_{i=1}^{N}\left[
    \left(
    \nabla \log p_{k}(s_i)
    \right)
    \left(
    \nabla \log p_{k}(s_i)
    \right)^T
\right],
\end{equation} 

However, calculating the FIM is computational expensive as the multiplication of $\nabla \log p_{k}(s_i)$ is 
both time and memory consumption when the size of the derivative matrix, i.e., $|\nabla \log p_{k}(s_i)|$, is substantial. Inspired by \cite{pascanu2013revisiting}, we calculate the diagonal of FIM to approximate the FIM as shown in Formula \ref{eq:appFIM}.
\begin{equation}
\label{eq:appFIM}
\tilde{\mathbf{F}}_{i} = I_{|\nabla \log p_{k}(s_i)|} \odot \mathbf{F}_{i}
\end{equation}
where $I_{|\nabla \log p_{k}(s_i)|}$ is the identity matrix with the same size of the derivative matrix. Then, we calculate the sum of the trace of $\tilde{\mathbf{F}}_{i}$ \cite{jastrzebski2021catastrophic} as the score of the data sample. Finally, we can calculate the difficulty score of a batch of data samples (see details in Appendix). 

In order to improve the training efficiency, we propose a curriculum data selection strategy. We take the simplest $\mathcal{B}^t_k$ data samples
for the local update on Device $k$ in Round $t$. $\mathcal{B}^t_k$ becomes bigger along with the epoch number within the training process (see details in Apendix).

\subsection{Efficient Sparse Parameter Update}

In this section, we propose an efficient sparse parameter method composed of a global aggregation layer selection method and a local update parameter selection method. 

\subsubsection{Global Aggregation Layer Selection}
\label{subsec:sparseLayerTraining}

In order to reduce communication costs, we only transfer the LoRA parameters in important layers (GAL) between the server and devices for global aggregation. In this section, we propose a global aggregation layer selection method with a novel layer importance score calculation technique and a global aggregation layer selection technique based on the importance score.

While important layers generally capture distinguishable features of data \cite{mellor2021neural}, we select the layers 
that are sensitive to the input data samples. When a layer exhibit less resilience against the noise on the data, it corresponds to higher sensitivity, and thus is more important. We calculate the output difference of a certain layer with two similar input data samples to indicate its resilience, which represents the importance score. 

In order to get two similar input data samples, we add noise to an original sample. Within a predefined noise budget, we calculate the noise that maximizes the loss, so as to well evaluate the sensitivity of the layer. Then, the noise ($\epsilon_i$) corresponding to $s_i$ is calculate based on Formula \ref{eq:noise}:
\begin{equation}
\label{eq:noise}
\begin{split}
     \epsilon_i = \argmax_{||\epsilon_i||_{p} < \gamma}  &~\underbrace{f(\mathcal{M}, \mathcal{P}_k, s_i + \epsilon_i, m_i)}_{L_{k}(s_i + \epsilon_i)} \\
     &~- \underbrace{f(\mathcal{M}, \mathcal{P}_k, s_i, m_i)}_{L_{k}(s_i)},
\end{split}
\end{equation}
where $L_k$ is the local loss, $||\cdot||_{p}$ represents the $\ell_p$-norm of the noise, and $\gamma$ refers to the noise budget. We decompose $L_{k}(s_i + \epsilon_i) - L_{k}(s_i)$ via the first-order Talyor extension as defined in Formula \ref{eq:firstOrder}:
\begin{equation}
\label{eq:firstOrder}
\begin{split}
    &~L_{k}(s_i + \epsilon_i) -L_{k}(s_i) \\
    \approx &~L_{k}(s_i) +     \epsilon_i^{T}\nabla_{\mathcal{P}_k} L_{k}(s_i) - L_{k}(s_i), \\
    = &~\epsilon_i^{T}\nabla_{\mathcal{P}_k}L_{k}(s_i).
\end{split}
\end{equation}
Then, we can solve the approximation by the solution to a classic dual problem \cite{foret2021sharpnessaware} as defined in Formula \ref{eq:fast_noise}.
\begin{equation}
\label{eq:fast_noise}
\begin{split}
    \epsilon_i^{*} &= \gamma\frac{ sign(\nabla_{\mathcal{P}_{k}}L_k(\mathcal{P}_{k}))|\nabla_{\mathcal{P}_{k}}L_k(\mathcal{P}_{k})|^{q-1}}{(||\nabla_{\mathcal{P}_{k}}L_k(\mathcal{P}_{k})||^{q}_{q})^{1/1-p}}
\end{split}
\end{equation}
where $|\cdot|^{q - 1}$ denotes the absolute value and power in terms of each element, $q$ is a factor that satisfies $\frac{1}{p} + \frac{1}{q} = 1$. Afterward, we take $\epsilon_i^{*}$ as the noise $\epsilon_i$ to calculate the sensitivity.

As Frobenius norm can characterize features in the latent space \cite{chen2021representation}, we exploit a relative difference of the Frobenius norm to measure the sensitivity of each layer. The relative difference can avoid the bias brought by the absolute values. The relative difference of of Frobenius norm is defined in Formula \ref{eq:frobenius}.
\begin{equation}
\label{eq:frobenius}
    \mathcal{F}^l(s_{i}) = \frac{||h^{l}(s_{i} + \epsilon^{*}_i)||_F - ||h^{l}(s_{i})||_F}{||h^{l}(s_{i})||_F},
\end{equation}
where $\mathcal{F}^l(s_{i})$ is the relative difference of the Frobenius norm, $h^{l}(s_{i})$ represents the output embeddings at Layer $l$ of the LLM $\mathcal{M}$ for data sample $s_{i}$, $||\cdot||_F$ refers to the Frobenius norm. Then, the importance score of Layer $l$ on Device $k$ is calculated based on its local data $D_k$ as defined in Formula \ref{eq:layerScoreDevice}.
\begin{equation}
\label{eq:layerScoreDevice}
    \mathcal{I}^l_k = \frac{1}{n_k} \sum_{s_i \in D_k} \mathcal{F}^l(s_{i}),
\end{equation}
where $\mathcal{I}^l_k$ represents the importance score of Layer $l$ on Device $k$. Afterward, the global importance score $\mathcal{I}^l$ is calculated based on Formula \ref{eq:layerScoreGlobal}.
\begin{equation}
\label{eq:layerScoreGlobal}
    \mathcal{I}^l = \frac{1}{N} \sum_{k = 1}^K n_k \mathcal{I}^l_k,
\end{equation}

We propose a lossless method to select proper important layers as GAL. On each device $k$, the LoRA parameters are initialized as $\mathcal{P}_{k}^0$. After $\mathcal{T}$ rounds of fine-tuning, the LoRA parameters are denoted by $\mathcal{P}_{k}^{\mathcal{T}}$. We construct a base function as $\Delta_{k} = \mathcal{P}_{k}^0 - \mathcal{P}_{k}^{\mathcal{T}}$. We calculate the Hessian matrix of the local loss function with its eigenvalues sorted in ascending order ($\{\lambda^1_{k}, \lambda^2_{k}, ..., \lambda^r_{k}\, ..., \lambda^{R_k}_{k}\}$ with $r$ representing the index of an eigenvalue and $R$ indicating the rank of the Hessian matrix). We calculate the Lipschitz constant ($\mathscr{L}_{k}$) of a base function $H_k(\mathcal{P}_{k}^{\mathcal{T}}) \Delta_{k} - \triangledown L'_k(\Delta_{k} + \mathcal{P}_{k}^{\mathcal{T}})$ with $\triangledown L'_k(\cdot)$ being the gradient of the local loss function and $H_k$ referring to the Hessian matrix of the local loss function. Inspired by \cite{zhang2021validating}, we find the first $r_{k}$ that satisfies $\lambda_{r_{k+1}} - \lambda_{r_{k}} > 4\mathscr{L}_{k}$ to achieve lossless performance. Then, we calculate the expected number of layers in GAL on Device $k$ as $\mathcal{N}^*_{k} = (1 - \frac{r_{k}}{R_k})L$ with $L$ being the number of layers in $\mathcal{M}$. Then, we calculate the number of layers in GAL as $\mathcal{N}^* = \frac{\mu}{N} \sum n_k \mathcal{N}^*_{k}$, where $\mu$ is a hyper-parameter to adjust the ratio between global and local number. Finally, we select $\mathcal{N}^*$ layers with the highest importance scores.

\subsubsection{Local Update Parameter Selection} 
\label{subsec:pupdate}

In order to reduce computation costs, we only update important LoRA parameters within the local update while freeze the remaining parameters. Apart from the parameters in GAL, we dynamically select an important part of parameters in other layers to update. In this section, we propose a novel fisher information-based local update parameter selection method with momentum. 


While the LoRA parameters may significantly vary during the fine-tuning process, we calculate the FIM with momentum within first $\mathcal{T}'$ epochs by $\mathbf{F}^t_k = \gamma*\mathbf{F}^{t-1}_k + (1-\gamma)\tilde{\mathbf{F}}_k$, where $\gamma$ represents the coefficient that controls the step size of the moving average, $\mathbf{F}^t_k$ refers to the FIM on Device $k$ at Round $t$, $\tilde{\mathbf{F}}_k$ is the empirical average diagonal approximation of the FIM, i.e., $\tilde{\mathbf{F}}_k = \frac{1}{n_k} \sum_{s_i \in D_k} \tilde{\mathbf{F}}_i $ with $\tilde{\mathbf{F}}_i$ calculated based on Formula \ref{eq:appFIM}, while $\mathbf{F}_k^0$ is directly calculated without moment. Finally, we get a FIM $\mathbf{F}^{\mathcal{T}'}_{k,l}$ for each layer $l$ outside of GAL. Inspired by \cite{Diao2023Pruning}, we exploit a neuron-wise aggregation of the FIM to indicate the importance score of Neuron $\mu$ in Layer $l$ as defined in Formula \ref{eq:neuron_sum}.
\begin{equation}
\label{eq:neuron_sum}
    \mathcal{s}_{k,l}^{\mu} = \sum_{\upsilon=0}^{|\mathcal{W}_{\mu:}|-1} \mathbf{F}_{k}^{\mathcal{T}'}[\mu * |\mathcal{W}_{\mu:}| + \upsilon]
\end{equation}
Where $|\mathcal{W}{\mu:}|$ denotes the number of elements in the $\mu^{th}$ row of the full weight matrix $\mathcal{W}_{o}^l$ in $\mathcal{M}$, $\mathbf{F}_{k,l}^{\mathcal{T}'}[\nu]$ represents the $\nu^{th}$ diagonal element in $\mathbf{F}_{k,l}^{\mathcal{T}'}$. Afterward, we exploit the lossless method to calculate the proper local update parameter ratio as $\rho_{k,l} = 1 - \frac{r_{k,l}}{R_{k,l}}$ (see details in Section \ref{subsec:sparseLayerTraining}, with $r_{k,l}$ and $R_{k,l}$ representing the corresponding $r_k$ and $R_k$ in Layer $l$. Finally, we take the most important $\rho_{k,l}$ neurons in terms of the importance score $\mathcal{s}_{k,l}^{\mu}$ as the local update parameters to be updated with the parameters in GAL and freeze the other parameters within the local update.

\subsection{\TheName{} Algorithm}

\begin{figure}[t]
\vspace{-4mm}
\begin{algorithm}[H]
\caption{Fisher Information-base Efficient Curriculum Federated Learning (\TheName{})}
\label{alg:ficfl}
\begin{algorithmic}[1]
\INPUT  \quad \newline
$T$: The maximum number of rounds \newline
$K$: The number of devices \newline
$D = \{D_1, D_2, ..., D_K\}$: The set of datasets on each device \newline
$\eta = \{\eta_1, \eta_2, ..., \eta_T\}$: The learning rates
\OUTPUT \quad \newline
$\mathbf{P}^{t}$: The set of LoRA parameters at Round $t$ 
\For{$k$ in $\{1, 2, ..., K\}$ (in parallel)} \label{line:initialBegin}
\For{$B_j \in D_k$} \label{line:scoreBegin}
\State $\mathcal{s}_j$ $\gets$ Calculation based on Formula 7
\EndFor \label{line:scoreEnd}
\State Sort $B_j \in D_k$ in ascending order of $\mathcal{s}_j$ \label{line:sort}
\EndFor
\State $GAL$ $\gets$ Compute the GAL \label{line:gal}
\For{$k$ in $\{1, 2, ..., K\}$ (in parallel)} \label{line:pupdateFor}
\State $\mathcal{P}^u_k$ $\gets$ Compute local update parameters \label{line:pupdate}
\EndFor \label{line:initialEnd}
\For{$t$ in $\{1, 2, ..., T\}$} \label{line:tuningBegin}
\State Sample $\mathcal{K} \subseteq \{1, 2, ..., K\}$ devices \label{line:deviceSampling}
\For{$k$ in $\mathcal{K}$} \label{line:localUpdateBegin}
\State Select $\mathcal{B}^t_k$ data samples based on Formula 8 \label{line:select}
\State $\mathcal{P}^{t - \frac{1}{2}}_k$ $\gets$ Update $\mathcal{P}^{t - 1}_k$ with $\mathcal{P}^{t - 1}_{GAL}$ \label{line:localSync}
\State $\mathcal{P}^t_k$ $\gets$ Update $\mathcal{P}^u_k \subset \mathcal{P}^{t - \frac{1}{2}}_k$ using $\mathcal{B}^t_k$ data samples \label{line:localUpdate}
\EndFor \label{line:localUpdateEnd}
\State $\mathcal{P}^t_{GAL}$ $\gets$ Aggregate $\mathcal{P}^t_{GAL,k}$ with $k \in \mathcal{K}$ \label{line:aggregation}
\EndFor \label{line:tuningEnd}
\end{algorithmic}
\end{algorithm}
\vspace{-6mm}
\end{figure}

The \TheName{} algorithm is shown in Algorithm \ref{alg:ficfl}. Within the initialization phase (Lines \ref{line:initialBegin} - \ref{line:initialEnd}), the difficulty scores of each batch are calculated based on Formula 7
(Lines \ref{line:scoreBegin} - \ref{line:scoreEnd}), the batches of data samples are sorted in ascending order in terms of the difficulty scores for the curriculum data selection strategy (Line \ref{line:sort}), the GAL are calculated in Line \ref{line:gal} (see details in Section 4.3.1
), and the local update parameters are computed in Line \ref{line:pupdate} (see details in Section 4.3.2
). Then, within the fine-tuning phase is performed in Lines \ref{line:tuningBegin} - \ref{line:tuningEnd}. A set of devices $\mathcal{K}$ is randomly selected (Line \ref{line:deviceSampling}). Then, on each device, local update is carried out in Lines \ref{line:localUpdateBegin} - \ref{line:localUpdateEnd}. First, the data samples are selected based on the curriculum data selection strategy in Line \ref{line:select}. Second, the LoRA parameters are updated with the global parameters in PSL transferred from the server in Line \ref{line:localSync} (see details in Section 4.1
). Third, the LoRA parameters are updated based on the local training with the selected $\mathcal{B}^t_k$ data samples in Line \ref{line:localUpdate} (see details in Section 4.3.2
). Finally, the parameters in the global aggregation layers are aggregated using the FedAvg algorithm 
(McMahan et al., 2017)
on the server (Line \ref{line:aggregation}).

While \TheName{} focuses on improving the efficiency of federated learning, it can further enhance the data privacy without bringing extra privacy issues. Within the training phase of \TheName{}, we only transfer the parameters in global aggregation layers between the server and devices, which can avoid transfer the whole model so as to protect the data privacy. In addition, we update the local update parameters instead of the full parameters, which can further avoid privacy and security issues due to potential gradient leakage. Traditional attack methods, e.g., gradient attack, assume that full gradients or models are transferred between the server and devices \cite{dimitrov2022dataleakagefederatedaveraging,Marchand2023SRATTA}. Thus, compared with the traditional federated learning approaches that transfer the whole model between the server and devices, \TheName{} can further enhance the privacy and security of federated learning.

\begin{table*}[ht!]
\large
\centering
\resizebox{0.9\linewidth}{!}{
\begin{tabular}{cccccccccccc}
\toprule
\textbf{Method}    & \textbf{QNLI}      & \textbf{SST-2}       & \textbf{CoLA}         & \textbf{MRPC}       & \textbf{RTE}       & \textbf{BoolQ}       & \textbf{MPQA}         & \textbf{Subj}       & \textbf{Trec}          & \textbf{MR}        & \textbf{Avg} \\
\midrule
Adapter                        & 49.46  & 90.83  & 54.17 & 84.77 & 47.29 & 62.17 & 90.95   & 51.65  & 96.2   & 91.30 & 58.32                 \\

FedPrompt                   & 87.73 & 94.38	& 19.79 & 76.31	& 64.98	& 74.58	&90.10	&94.25	&92.6	&91 & 78.57               \\
P-tuning v2                    & 88.74	& 94.04 &50.23	&78.16	&76.17	&74.89	&88.75	& 95.5 & 90.8	&90.65 & 82.79                 \\

IDPG               & 66.7	&89.11	&4.59	&72.22	&52.35	&68.93	&71.8	&59.4	&73.4	&86.35 & 64.48                 \\
ATTEMPT                &50.74	&50.92	&4.63	&76.01	&54.15	&62.17	&90.35	&88.85	&77.2	&91.15& 64.16                \\
LPT                      & 89.38	& 94.27	&50.78	&82.38	&80.86	&62.2	&90.15	&95.75	&92.4	&90.6& 82.87      \\

LORA                   & 89.86	&94.72	&54.78	&83.15	&78.7	&75.96	&89.2	&95.8	&94.4	&91.35 &84.72                 \\   
Shortformer                   & 90.17	&94.04	&54.16	&84.49	&79.78	&78.62	&90.6	&96.85	&95.2	&91.1 & 85.5\\ 

VOC                 & 91.89	&95.18	&53.64	&85.15	&81.31	&78.32	&90.85	&\underline{96.85}	&96.6	&91.25 & 86.1\\ 

SLW                   & 91.91	&94.1	&54.75	&85.01	&79.06	&78.41	&91.25	&96.6	&95.8	&91.15& 85.80\\ 

PFedGate                  & 90.72	& 93.81	& 50.73	&83.33	&76.17	&76.88	&89.8	&94.5	&94	&87.65& 83.75\\ 

FedDST                  & 90.15	&94.61	&30.12	&81.41	&71.84	&77.13	&90.05	&95.85	&86.2	&91.2& 80.85\\ 

SE                  & 76.42	&93.69	&55.47	&81.72	&71.84	&76.42	&89.8	&96.3	&87.2	&91.85& 82.07\\

FedALT                   & 91.76	&94.15	&\underline{55.5}	&\underline{85.89}	&80.95	&80.46	&90	&96.15	&96.6	&91.25& 86.23\\  

sLORA                  & 92.40	&\underline{94.38}	&55.51	&85.41	&82.31	&80.24	&\underline{91.30}	&96.8	&\underline{97.2}	&91.35& \underline{86.69}\\ 

adaLORA                   & 91.14	&92.32	&44.46	&81.83	&75.81	&77.16	&89.65	&96.25	&91	&\underline{91.8}& 83.14\\ 

Delta-LoRA                   & \underline{92.41}	&94.27	&54.95	&85.36	&\underline{83.39}	&\underline{80.18}	&90.80	&96.75	&96.8	&91.25& 86.61\\
\hdashline
Ours                   & \textbf{93.12}	&\textbf{95.76}	&\textbf{58.57}	&\textbf{90.85}	&\textbf{84.69}	&\textbf{80.92}	&\textbf{91.35}	&\textbf{97.0}	&\textbf{97.8}	&\textbf{92.95}& \textbf{88.31}\\ 
\bottomrule
\end{tabular}
}
\vspace{-2mm}
\caption{The convergence accuracy with \TheName{} and diverse baseline approaches. The evaluation with GLUE benchmark is based on development sets while others are based on test sets. The best results are highlighted in \textbf{bold} and the second bests are marked with \underline{underline}. The results are obtained using RoBERTa\textsubscript{LARGE}.}
\vspace{-6mm}
\label{tab:acc_results_roberta}
\end{table*}

\section{Experiments}
\label{sec:exp}

In this section, we demonstrate extensive experimentation with 17 baseline approaches and 10 NLP tasks to
reveal the advantages of \TheName{}.

\subsection{Experimental Setup}
We take an FL environment composed of 100 devices and a parameter server in the experimentation. We randomly sample 10 devices in each epoch. We utilize 10 \rev{c}ommonly-used NLP tasks in the experimentation, i.e., QNLI \cite{Rajpurkar2016SquAD}, SST-2 \cite{Socher2013Recursive}, CoLA \cite{Warstadt2019Neural}, MRPC \cite{DolanB2005Automatically}, RTE \cite{Giampiccolo2007The}, BoolQ \cite{Clark2019Boolq}, MPQA \cite{Wiebe2005Annotating}, Subj \cite{Pang2004A}, TREC \cite{Voorhees2000Building}, and MR \cite{Pang2005Seeing}. The input data of the tasks are non-IID among the 100 devices. We compare \TheName{} with 17 baseline approaches, i.e., a parameter efficient fine-tuning-based approaches (Adapter \cite{Houlsby2019Parameter}), 6 prompt-based tuning methods (FedPrompt \cite{Zhao2023FedPrompt}, P-tuning v2 \cite{liu2022p}, IDPG \cite{Wu2022IDPG}, ATTEMPT \cite{Asai2022ATTEMPT}, LPT \cite{Liu2022Late}, LoRA \cite{hu2021lora}), 4 curriculum learning-based approaches (Shortformer \cite{Press2020aShortformer}, VOC \cite{platanios2019Competence}, SLW \cite{li2024deepspeed}, SE \cite{Peng2023Token}), 3 personalized FL methods (PFedGate \cite{Chen2023Efficient}, FedDST \cite{Bibikar2022Federated}, FedALT \cite{Pillutla2022Federated}), and 2 Lora based methods (SLoRA \cite{Babakniya2023SLoRA}, AdaLoRA \cite{zhang2023adaptive}). 
We carry out the experimentation based on two language models, i.e., RoBERTa\textsubscript{LARGE} \cite{Liu2020RoBERTa} and LLaMA \cite{touvron2023llama}. 

\subsection{Evaluation of \TheName{}}

In this section, we present the evaluation of \TheName{} based on RoBERTa\textsubscript{LARGE} and LLaMA.

\subsection{Evaluation based on RoBERTa\textsubscript{LARGE}} 

Table \ref{tab:acc_results_roberta} present the convergence accuracy of diverse approaches based on RoBERTa\textsubscript{LARGE}. \TheName{} significantly outperforms baseline methods in terms of the convergence accuracy (up to 45.35\%, 38.37\%, 12.69\%, 37.60\%, 42.38\%, 18.72\%, 7.01\%, 6.36\%, 5.70\%, 5.90\%, 8.79\%, 28.45\%, 16.70\%, 4.96\%, 5.44\%, 14.11\% and 5.49\% compared with Adapter, FedPrompt, P-tuning v2, IDPG, ATTEMPT, LTP, LoRA, Shortformer, VOC, SLW, PFedGate, FedDST, SE, FedALT, sLoRA, AdaLoRA, and Delta-LoRA respectively). In addition, we analyze the time to achieve target accuracy (see details in Appendix), which demonstrates that \TheName{} outperforms baseline methods in terms of efficiency (up to 94.7\%, 97.43\%, 98.61\%, 61.64\%, 96.12\%, 91.6\%, 96.69\%, 89.12\%, 80.15\%, 84.26\%, 80.82\%, 97.01\%, 96.52\%, 82.16\%, 85.18\%, 95.66\% and 69.65\% compared with Adapter, FedPrompt, P-tuning v2, IDPG, ATTEMPT, LTP, LoRA, Shortformer, VOC, SLW, PFedGate, FedDST, SE, FedALT, sLoRA, AdaLoRA, and Delta-LoRA respectively). The advantages are expected as  our proposed curriculum data selection strategy on each device can well improve both the efficiency and the effectiveness. In addition, the proposed important layer selection method can reduce the scale of parameters to transfer between the server and devices. Furthermore, the local update parameter selection method can well reduce useless computation on each devices while freezing unimportant parameters may mitigate the effect of overfitting brought by the non-IID data.

\subsection{Evaluation based on LLaMA 7B}

\begin{table}[t!]
\resizebox{\linewidth}{!}
{
\begin{tabular}{lllllll}
    \toprule
    \multirow{2}{*}{\textbf{Method}} & \multicolumn{2}{c}{\textbf{COLA}}  & \multicolumn{2}{c}{\textbf{MRPC}} & \multicolumn{2}{c}{\textbf{RTE}} \\
    \cmidrule(r){2-7}   & \textbf{Acc} & \textbf{Time} & \textbf{Acc} & \textbf{Time} & \textbf{Acc} & \textbf{Time}    \\
    \midrule
    FedPrompt           & 59.30 & 2527 & 80.54 & 1365 & 56.68 & 1296\\
    P-tuning v2         & 3.89 & 2159& 80.18 & 935 & 52.35 & 842 \\
    ATTEMPT             & 54.77 & 1825 & 81.38 & 1069 & 74.37 &  934\\
    LPT                & 51.8 & 1645& 80.13 & 958 & 74.01 & 1045 \\
    LoRA                & 59.55 &1768& 79.56 &941&80.87&984 \\
    Voc         & 60.64 &1364&38.64 & 859& 75.28 &745\\
    SE         & 58.69 & 1628 & 79.35 &846& 74.01&839 \\
    SLoRA         & 60.56 &1602 & 80.39 &947 & 59.93 &945\\
    DeltaLoRA         & 58.91 &1674& 81.67 &1034& 81.95& 1329 \\
    \hdashline
    \TheName{}          & \textbf{61.48} & \textbf{1298}& \textbf{81.93}& \textbf{832}& \textbf{86.28} & \textbf{703} \\
    \bottomrule
\end{tabular}
}
\vspace{-2mm}
\caption{Convergence accuracy and fine-tuning time on COLA, MRPC and RTE with LLaMA.}
\vspace{-4mm}
\label{tab:results_llama}
\end{table}






We carried out the experimentation with a LLM, i.e., LLaMA 7B on MRPC, MR, and SST-2 dataset. As shown in Table \ref{tab:results_llama}, \TheName{} significantly outperforms baseline approaches in terms of both performance (up to 29.60\%, 33.93\%, 11.91\%, 12.27\%, 5.41\%, 11\%, 12.27\%, 26.35\%, 4.69\% higher accuracy compared with FedPrompt, P-tuning v2, ATTEMPT, LPT, LORA, VOC, SE, SLoRA and DeltaLoRA) and efficiency (up to 45.75\%, 39.87\%, 28.8\%, 32.7\%, 26.58\%, 5.63\%, 16.2\%, 25.6\%, 47.1\% faster compared with FedPrompt, P-tuning v2, ATTEMPT, LPT, LORA, VOC, SE, SLoRA and DeltaLoRA). The advantages reveal the our proposed approach improves both the efficiency and effectiveness with LLM.

\subsection{Robustness \& Scalability}

In this section, we demonstrate the robustness and the scalability of \TheName{} with divers degrees of non-IID data and different device numbers. \TheName{} achieves comparable accuracy across divers degrees of data heterogeneity, the difference of which is smaller than 1.83\%. In addition, we conduct the experimentation with the device number ranging from 20 to 100 based on MRPC dataset and find that the disparity is smaller than 0.79\% in terms of accuracy. 

\subsection{Communication overhead}

The absolute communication overhead is shown in the Table \ref{tab:abcom} of Appendix (with RoBERTa\textsubscript{LARGE}). The communication overhead of FibecFed is higher than that of FedPrompt (up to 3.51 times), IDPG (up to 3.3 times), and ATTEMPT (up to 1.85 times). This is expected as the these three methods are prompt tuning-based methods, which corresponds to much fewer parameters to update during the training phase compared with FibecFed. However, these three methods correspond to significantly lower performance (compared with FibecFed), i.e., low convergence accuracy (from 1.25\% to 38.68\% for FedPrompt, from 6.6\% to 53.98\% for IDPG, and from 1\% to 53.94\% for ATTEMPT), as shown in Table 1. The communication overhead of FibecFed is significantly lower than the other methods (6.25 times for Adapter, 9.67 times for P-tuning V2, 1.9 times for LPT, and 25\% for LORA, SHORTFORMER, Voc, SLW, PFedGate, FedDST, SE, FedAlt, sLora, AdaLora, Delta-LoRA). This is expected as well as FibecFed only transfers the global aggregation layers instead of the parameters of all the layers in LLM. 

In addition, the relative communication overhead (i.e., the ratio between the absolute communication overhead and the total training time) is shown in Table \ref{tab:recom} of Appendix. Similar to the absolute communication overhead, the relative communication overhead of FibecFed is higher than that of FedPrompt (from 2.3\% to 37.4\%), IDPG (from 2.4\% to 34.4\%), and ATTEMPT (from 1.9\% to 31.7\%) as well. This is expected as explained before. As the total training time of FibecFed becomes shorter, the relative communication overhead of FibecFed becomes slightly more significant than FedAlt (from 1.4\% to 20.1\%) and AdaLora (from 0.8\% to 5.7\%). In addition, the relative communication overhead of FibecFed becomes similar to that of sLora (from 7.8\% smaller to 24.8\% bigger) and Delta-LoRA (from 7.5\% smaller to 5.5\% bigger).  The relative communication overhead of FibecFed is still smaller than the rest approaches, i.e., Adapter (up to 43.9\%), P-tuning V2 (up to 55.6\%), LPT (up to 25.1\%), LORA (up to 14.5\%), SHORTFORMER (up to 13.8\%), Voc (up to 8.0\%), SLW (up to 8.8\%), PFedGate (up to 7.0\%), FedDST (up to 7.1\%), SE (8.1\%). Please note that FibecFed corresponds to higher convergence accuracy (as shown in Table \ref{tab:acc_results_roberta}) and shorter training time (as shown in Table \ref{tab:results_llama}).

\subsection{Ablation Study}

To analyze the impact of each module in \TheName{}, we demonstrate the ablation study in terms of the curriculum data selection method, the important layer selection method, and the local update parameter selection method. First, we conduct a comparative analysis among four curriculum strategies, i.e., SLW, VOC, Shortformer, SE, and that without curriculum (NULL). \TheName{} corresponds to superior performance in terms of accuracy (up to 5.73\%, 9.12\% 5.84\%, 6.41\%, 7.7\% compared with Voc, SE, SLW, Shortformer, and NULL, respectively) and efficiency (up to 26.53\%, 34.26\% 68.92\%, 68.36\%, 58.57\% compared with Voc, SE, SLW, Shortformer, and NULL, respectively). Afterward, we compare the importance layer selection method with Ascending Order (AO), Descending Order, Random Order (RO) and that with full layer synchronization (FULL), which reveals the advantages of our layer selection method in terms of accuracy (up to 3.42\%, 2.76\%, 2.65\%, 1.02\%) and efficiency (up to 29.3\%, 18.46\%, 23.1\%, 15.3\%) compared with AO, DO, RO, and FULL, respectively. Furthermore, we compare our local update parameter selection method to that without selection, i.e., all the parameters are updated. The advantage of our local update parameter selection method can achieve 2.48\% higher accuracy and 11.8\% faster.

\section{Conclusion}
\label{sec:con}

In this paper, we propose an original fisher information-based efficient curriculum federated learning, i.e., \TheName{}. Within \TheName{}, we propose an adaptive federated curriculum learning method and an efficient sparse parameter update method. We exploit fisher information to calculate the difficulty scores of data samples and propose the an original curriculum data selection strategy. In the sparse parameter update method, we propose a new sensitivity-based important layer selection technique and a novel fisher information-based important parameter technique method while freezing the remaining parameters to achieve both efficiency and effectiveness. We demonstrate the results of extensive experimentation to compare \TheName{} with 17 baseline approaches based on 10 NLP tasks, which reveal significant advantages of \TheName{} in terms of accuracy (up to 45.35\%) and fine-tuning speed (up to 98.61\% faster).



\section*{Limitations}

\rev{While our approach can be to exploited to significantly improve the performance and efficiency of LLM federated learning, we assume that a central parameter server is exploited to coordinate the training process. When there is not a central parameter or the heterogeneous devices \cite{che2022federated,li2022fedhisyn} are connected based on diverse topologies (decentralized setting \cite{liu2024aedfl}), e.g., ring, it would be complicated to directly exploit our proposed approach. In addition, our approach can be combined with model compression methods, e.g., model pruning and quantization \cite{jia2024efficient}, to achieve better performance. In the future, we anticipate exploiting model compression methods in LLM federated learning with a decentralized setting.}

\bibliography{sample.bib}

\clearpage

\appendix

\section{FedAvg Update}

The original FedAvg update is shown in Algorithm \ref{alg:fedavg}.

\begin{figure}[t!]
\vspace{-4mm}
\begin{algorithm}[H]
\caption{FedAvg}
\label{alg:fedavg}
\begin{algorithmic}[1]
\INPUT  \quad \newline
$t$: The number of current round \newline
$\mathcal{P}^{t-1}$: The global LoRA parameters at Round $t-1$ \newline
$T$: The maximum number of rounds \newline
$K$: The number of devices \newline
$D = \{D_1, D_2, ..., D_K\}$: The set of datasets on each device \newline
$\lambda = \{\lambda_1, \lambda_2, ..., \lambda_T\}$: The learning rates
\OUTPUT \quad \newline
$\mathcal{P}^{t}$: The global LoRA parameters at Round $t$ 
\For{$k$ in $\{1, 2, ..., K\}$ (in parallel)}
\State $\mathcal{P}_{k} \gets \mathcal{P}^{t-1}$
\For{$B_{j} \in D_{k}$}
\State $\mathcal{P}_k$ $\gets$ $\mathcal{P}_{k}$ -  $\lambda^t \sum_{s_i \in B_{j}} \nabla_{\mathcal{P}_k} f(\mathcal{M}, \mathcal{P}_k, s_i, m_i)$
\EndFor
\EndFor
\State $m^{t} \gets \sum_{k=1}^{K} |D_{k}|$
\State $\mathcal{P}^{t} = \sum_{k=1}^{K}\frac{|D_k|}{m^{t}}\mathcal{P}_{k}$
\end{algorithmic}
\end{algorithm}
\vspace{-10mm}
\end{figure}

\section{Gradient Calculation in LoRA}

During the local training, we update LoRA parameters at Epoch $t$ for data samples $s_i$ as follows:
\begin{equation}
\begin{split}
    ~ \mathcal{P}_{A}^{t} &= \mathcal{P}_{A}^{t-1} - \lambda^t \nabla_{\mathcal{P}_{A}^{t-1}} f(\mathcal{M}, \mathcal{P}_k, s_i, m_i)\\
    ~ \mathcal{P}_{B}^{t} &= \mathcal{P}_{B}^{t-1} - \lambda^t \nabla_{\mathcal{P}_{B}^{t-1}} f(\mathcal{M}, \mathcal{P}_k, s_i, m_i),\\
\end{split}
\end{equation}
and for a batch of data samples $B_j$ as follows:
\begin{equation}
\begin{split}
    ~ \mathcal{P}_{A}^{t} &= \mathcal{P}_{A}^{t-1} - \lambda^t \sum_{s_i \in B_{j}} \nabla_{\mathcal{P}_{A}^{t-1}} f(\mathcal{M}, \mathcal{P}_k, s_i, m_i)\\
    ~ \mathcal{P}_{B}^{t} &= \mathcal{P}_{B}^{t-1} - \lambda^t \sum_{s_i \in B_{j}} \nabla_{\mathcal{P}_{B}^{t-1}} f(\mathcal{M}, \mathcal{P}_k, s_i, m_i),\\
\end{split}
\end{equation}
where $\lambda^t$ is the learning rate, $\mathcal{P}_k$ is the combination of $\mathcal{P}_A$ and $\mathcal{P}_B$. To get the gradient of reconstructed full-rank matrix at iteration $t$, we can calculate as follows:
\begin{equation}
    \nabla_{\mathcal{P}_{k}^t} f(\mathcal{M}, \mathcal{P}_k, s_i, m_i) = \mathcal{P}_{A}^{t} \mathcal{P}_{B}^{t} - \mathcal{P}_{A}^{t-1} \mathcal{P}_{B}^{t-1}
\end{equation}

\section{Formulas for Curriculum Federated Learning}

We can calculate the difficulty score of a data sample as defined in Formula \ref{eq:score_sample}.
\begin{equation}
\label{eq:score_sample}
    \mathcal{s}_i = Tr(\tilde{\mathbf{F}}_{i}),
\end{equation}
where $\mathcal{s}_i$ is the difficulty score of data sample $s_i$. Then, the difficulty score of a batch of data samples can be calculated based on Formula \ref{eq:score_batch}:
\begin{equation}
\label{eq:score_batch}
    \mathcal{s}_j = \sum_{s_i \in B_j} Tr(\tilde{\mathbf{F}}_{i}),
\end{equation}
where $\mathcal{s}_j$ is the difficulty score of the batch $B_j$. The more significant the score is, the more difficulty the batch of data samples is. We calculate the difficulty score based on the initial model as the difficulty of data samples corresponds to negligible change during the fine-tuning process \cite{platanios2019Competence,li2022the}. $\mathcal{s}_i$ can be computed using the square of the elements in the diagonal of the first-order derivative matrix, which can avoid the computation of the full FIM with the heavy multiplication of two matrices. This mechanism makes the calculation feasible in terms of computation time and memory consumption. Please note that FIM is calculated and stored locally, which does not need to be transferred to the server.

We calculate $\mathcal{B}^t_k$ based on Formula \ref{eq:cl}.
\begin{equation}
\label{eq:cl}
    \mathcal{B}^t_k = (\beta + (1 - \beta)\frac{t}{\alpha T})\frac{n_k}{\mathrm{B}},
\end{equation}
where $\beta$ represents the initial training sample ratio, $\alpha$ denotes the ratio of training epoch until all data is used, $\mathrm{B}$ refers to the batch size. Both $\beta$ and $\alpha$ are hyper-parameters within the range of $[0,1]$.
Then, the batch of data samples is selected based on Formula \ref{eq:select}:
\begin{equation}
\label{eq:select}
    Select^t(B_{j}) = 
    \begin{cases}
        True&  \text{if } j < \mathcal{B}^t_k \\
        False& \text{otherwise},
    \end{cases}
\end{equation}
where $Select^t(B_{j})$ represents the selection decision. When $Select^t(B_{j}) = True$, the batch of data samples is selected for local update. With this curriculum training strategy, the local training can learn from easy samples to challenging samples, which can achieve excellent performance with fewer data samples in early iterations. 

\section{Novelty of \TheName{}}

Different from the existing approaches, we exploit the fisher information to measure both the complexity of training data and the importance of the components, e.g., neurons, in LLMs. We enable curriculum learning based on the complexity of the training data within each device to achieve superb accuracy. We exploit LoRA to achieve efficient fine-tuning with only a few parameters. In addition, we split the trainable parameters into three parts, each of which is for global aggregation, local update, and as frozen neurons on each device, respectively. Only the parameters of the global aggregation part are updated and synchronized among multiple devices during the fine-tuning process. We consider the relative change of each layer and get an excellent estimation of important layers to generate the part for global aggregation. We exploit the momentum of parameter updating and get a robust estimation of important neurons in order to select important neurons for local update.

\section{Prompt-tuning in LLM}

When exploiting the prompt tuning, the parameters of the networks to generate prompts or prefix are adjusted instead of the parameters of LLMs. The prompt or the prefix is the added instruction concatenated to the input, which can guide LLMs to generate proper answers. For instance, a prompt (``This is [MASK]'') can be concatenated to the input (``Wonderful movie!'') to be sent to a LLM, which generates the label (``positive'' or ``negative'') for a sentiment analysis task. 

\section{Notations}

In this paper, we use the notations summarized in Table \ref{tab:summary}. 

\begin{table*}[t]
\caption{Summary of main notations.}
\label{tab:summary}
\begin{center}
\begin{tabular}{cc}
\hline
Notation & Definition \\
\hline
$\mathcal{M}$; $\mathbf{P}$ & The set of LLM parameters; the set of trainable LoRA parameters\\
$\mathcal{K}$; $K$ & The set of edge devices; the size of $\mathcal{M}$ \\
$\mathcal{D}$; $N$ & The global dataset; the size of $\mathcal{D}$\\
$\mathcal{D}_k$; $N_{k}$ &  The dataset on Device $k$; the size of $\mathcal{D}_k$\\
$\mathcal{F}(\cdot)$; $F_k(\cdot)$ & The global loss function; the local loss function on Device $k$ \\
$s_{i}$; $m_{i}$ & The single data sample; it's corresponding label of index $i$ \\

$L$ & The number of Layers\\
$T$ & The maximum number of global rounds \\
$\eta_{k}$& The learning rate on device $k$ \\
$\mathcal{K}$ & The number of participated devices per round  \\
$\mathcal{P}_{k}$ & The set of trainable LoRA parameters for device $k$\\
$W_{0}^{l}$ & The parameters at Layer $l$\\
$A_{k}^{l}, B_{k}^{l}$ & The LoRA matrices at Layer $l$ on device $k$\\
$h^{l}$ & The hidden values generated at Layer $l$\\
$\tilde{\mathbf{F}}_{i}$; $\mathbf{F}_{i}$ & The empirical and approximated FIM\\
$\odot$ & The hadamard product operation\\
$I$ & The identity matrices\\
$Tr$ & The trace of the matrices\\
$\mathcal{s}_i$;$ \mathcal{s}_j$ & The difficulty score of Sample $i$ and Batch $j$\\
$\epsilon_i$ & The generated noise for sample $i$\\
$p$;$q$ & The dual norm factors\\
$\gamma$ & The noise budget\\
$\mathcal{F}^l(s_{i})$ & The relative difference of Frobenius norm\\
$ \mathcal{I}^l$ & The importance score of Layer $l$\\
$\lambda_{k}$ & The $ith$ eigenvalues \\
$\mathscr{L}_{k}$ & The Lipschitz constant \\
$H_k(\mathcal{P}_{k}^{\mathcal{T}})$ & The Hessian Matrix with respect to $\mathcal{P}_{k}$ at round $\mathcal{T}$ on device $k$\\
$\mathbf{F}^t_k$ & The FIM on Device k at Round t\\
$\tilde{\mathbf{F}}_k^{t}$ & The empirical average diagonal approximation of $\mathbf{F}^t_k$\\
$\mathbf{F}_{k,l}^{\mathcal{T}'}$ & The FIM on Device k at Round t for layer $l$\\ 
$\mathbf{F}_{k,l}^{\mathcal{T}'}[\nu]$ & The $v^{th}$ diagonal element in $\mathbf{F}_{k,l}^{\mathcal{T}'}$\\
$R$;$r$ & The Rank of the Hessian Matrix; the index of eigenvalue.\\
$\mathcal{P}_{A}^{t}$;$\mathcal{P}_{B}^{t}$ & The fist and second part of LoRA parameters for device $k$\\
\hline
\end{tabular}
\end{center}
\end{table*}

\section{Experiment results}
In this section, we present the details of extensive experiments. In Section \ref{subsec:exp_setup}, we explain the details of experimental setup. In Section \ref{subsec:tuning_efficiency}, we compare the efficiency in terms of time to achieve target accuracy. In Section \ref{subsec:conver_curves}, we show the convergence accuracy over 10 different datasets. In Section\ref{subsec:robustness}, we present the robustness of \TheName{} respect to scalability and data heterogeneity. In Section \ref{subsec:lr}, we demonstrate the performance with learning rates. In Section \ref{subsec:CL_stra}, we discuss the impact of different curriculum strategies.

\subsection{Experimental Setup}\label{subsec:exp_setup}

We present the hyper-parameters used in the fine-tuning process in Table \ref{tab:exp_setup}. We set the global training epochs to 100, except for QNLI, SST-2. We follow the Dirichlet distribution (with 1 as concentration $\alpha$) to partition the whole data into splits and assign a certain number of samples based on Dirichlet distribution (with $\alpha=5$), development sets are served as test data to evaluate performance in the GLUE benchmark. For 4 other datasets, we select a certain number of samples from the training set as the development set, and the number of samples for each label is determined according to its original label distribution of training set. For datasets in GLUE benchmark, we use their original data splits. For 4 other datasets with no default splits, we randomly split the dataset into train, development, and test sets. 

With 100 devices, the number of samples ranges from 346 to 2607 for QNLI, 222 to 1676 for SST-2, 28 to 213 for COLA, 12 to 91 for MRPC, 8 to 62 for RTE, 31 to 235 for BoolQ, 25 to 189 for MPQA, 23 to 174 for Subj, 16 to 123 for Trec and 25 to 191 for MR. Additionally, a detailed number of samples distributed among 10 devices with the MRPC dataset is shown in Table \ref{tab:datasetDevice}. 

\begin{table*}[h]
\centering
\resizebox{0.75\linewidth}{!}{
\begin{tabular}{ccccccccccc}
\toprule
Device index    &  1  & 2  & 3  & 4  & 5  & 6  & 7  & 8  & 9  & 10    \\

\midrule

Number of samples & 422 & 317 & 303 & 303 & 651 & 474 & 270 & 431 & 378 & 119 \\

\bottomrule
\end{tabular}
}
\caption{The number of samples on each client for MRPC dataset.}
\label{tab:datasetDevice}
\end{table*}

RoBERTa\textsubscript{LARGE} consists of 24 layers of transformers followed by a classification head, which contains 355M parameters. LLaMA is composed of 32 transformer layers with 7B parameters. 

The centralized methods (Adapter, P-tuning v2, IDPG, ATTEMPT, LPT, LoRA, Shortformer, VOC, SLW, AdaLoRA, Delta-LoRA) are adapted to the FL setting with FedAvg \cite{mcmahan2017communication} for a fair comparison. In addition, we exploit the LoRA with FL opmization-based methods, i.e., PFedGate, FedDst and FedALT.

\subsection{Comparison with Random Data Selection}

The heuristic based methods or mode-oriented methods cannot address the heterogeneity issue across local data samples with variance in such metric. To show the effectiveness of our proposed Fisher information-based metric, we implement another baseline method using random selecting in the curriculum data selection strategy. Table \ref{tab:accRandom} shows the performance under different selecting strategies. Compared with other selecting method, FibecFed outperforms other baselines up to 8.51\% in terms of accuracy, which demonstrates the effectiveness of proposed method. In addition, Random corresponds to the lowest accuracy compared other methods, i.e., ShortFormer, SLW, Voc, and FibecFed. Table \ref{tab:randomTime} also indicates that FibecFed achieves the target accuracy of 85\% on the MRPC dataset in less time(up to 92.49\% faster) compared with other selection strategies. It is noteworthy that the random method reaches the accuracy 85\% at epoch 95, significantly increasing the required time.
This will be clarified in the revised version.

\begin{table*}[h]
\centering
\resizebox{0.26\linewidth}{!}{
\begin{tabular}{cc}
\toprule
Method    & Accuracy    \\

\midrule

ShortFormer   & 86.05\%    \\
SLW        &   86.58\%      \\
Voc        &   86.49\%      \\
Random        &   85.56\%      \\
FibecFed        &   90.84\%      \\
\bottomrule
\end{tabular}
}
\caption{The accuracy under different sample selection strategies with RoBERTA-Large model on mrpc dataset.}
\label{tab:accRandom}
\end{table*}

\begin{table*}[h]
\centering
\resizebox{0.21\linewidth}{!}{
\begin{tabular}{cc}
\toprule
Method    & Time    \\

\midrule

ShortFormer   & 173    \\
SLW        &   177     \\
Voc        &   118    \\
Random        &   786.6     \\
FibecFed        &   59      \\
\bottomrule
\end{tabular}
}
\caption{The time(seconds) under different sample selection strategies to achieve target accuracy 85\% with RoBERTA-Large model on mrpc dataset.}
\label{tab:randomTime}
\end{table*}

\subsection{Fine-tuning Efficiency}\label{subsec:tuning_efficiency}

In order to show the efficiency of \TheName{}, we present the time to achieve target accuracy in Table \ref{tab:time_roberta}.

\subsection{Accuracy \& Time Figures in RoBERTa\textsubscript{LARGE}}\label{subsec:conver_curves}

\begin{figure*}[t]
\centering

\subfigure[Acc \& COLA]{
\includegraphics[width=0.31\textwidth]{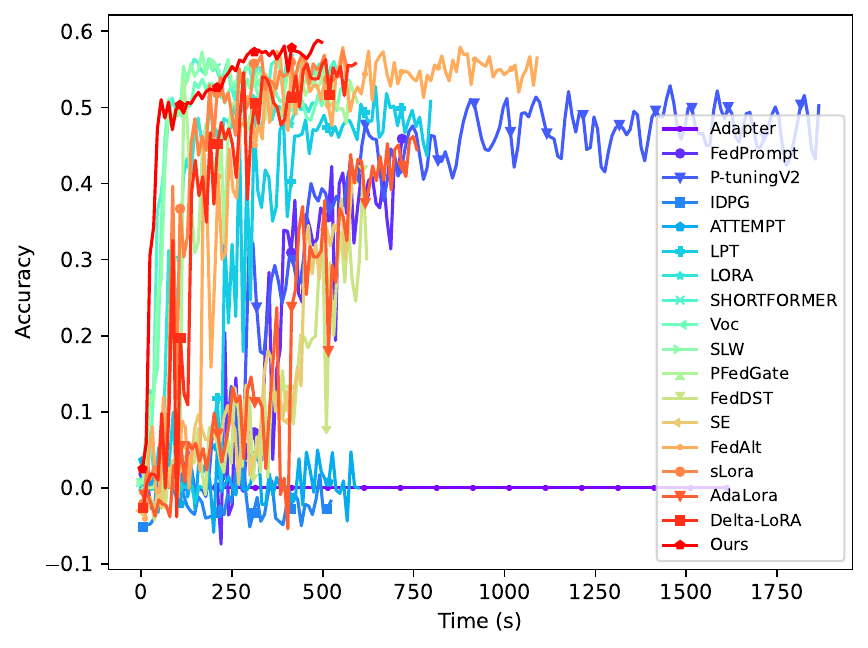}
\label{fig:cola}
}
\subfigure[Acc \& QNLI]{
\includegraphics[width=0.31\textwidth]{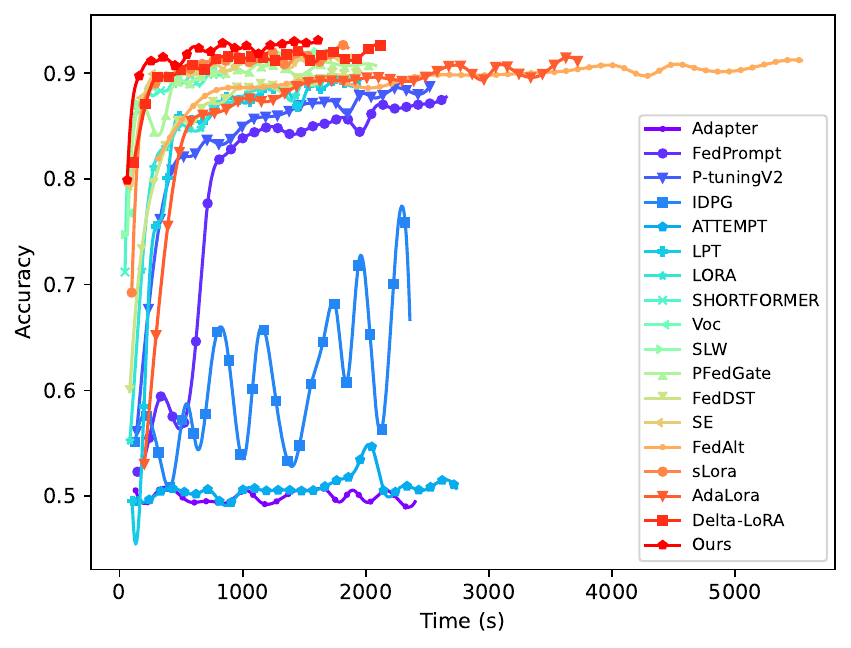}
\label{fig:qnli}
}
\subfigure[Acc \& SST-2]{
\includegraphics[width=0.31\textwidth]{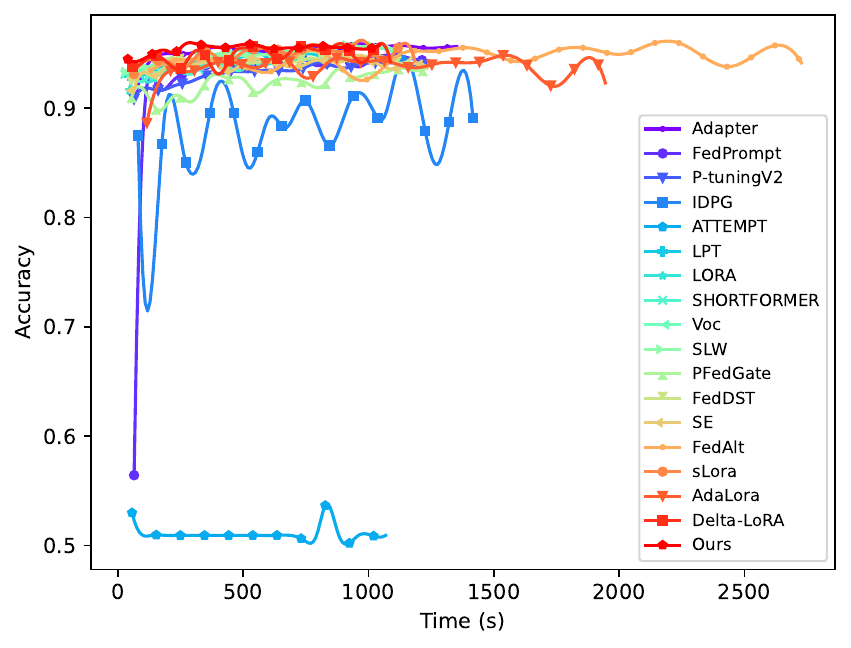}
\label{fig:sst-2}
} 
\vspace*{-3mm}
\caption{The accuracy and training time with \TheName{} and diverse baseline approaches.}
\vspace*{-2mm}
\label{fig:cola_qnli_sst2}
\end{figure*}

\begin{figure*}[h]
\centering

\subfigure[Acc \& MRPC]{
\includegraphics[width=0.31\textwidth]{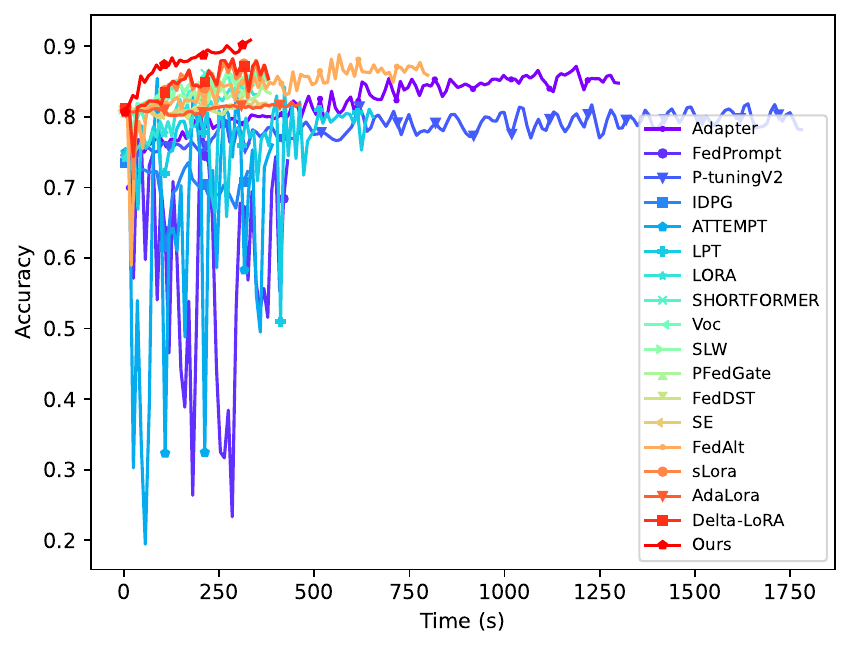}
\label{fig:mrpc}
}
\subfigure[Acc \& RTE]{
\includegraphics[width=0.31\textwidth]{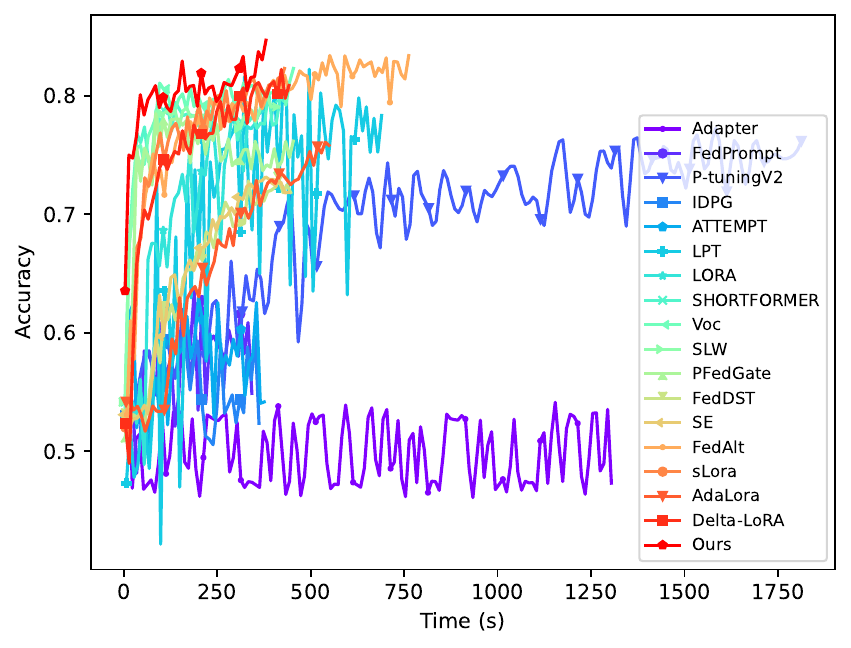}
\label{fig:rte}
}
\subfigure[Acc \& BOOLQ]{
\includegraphics[width=0.31\textwidth]{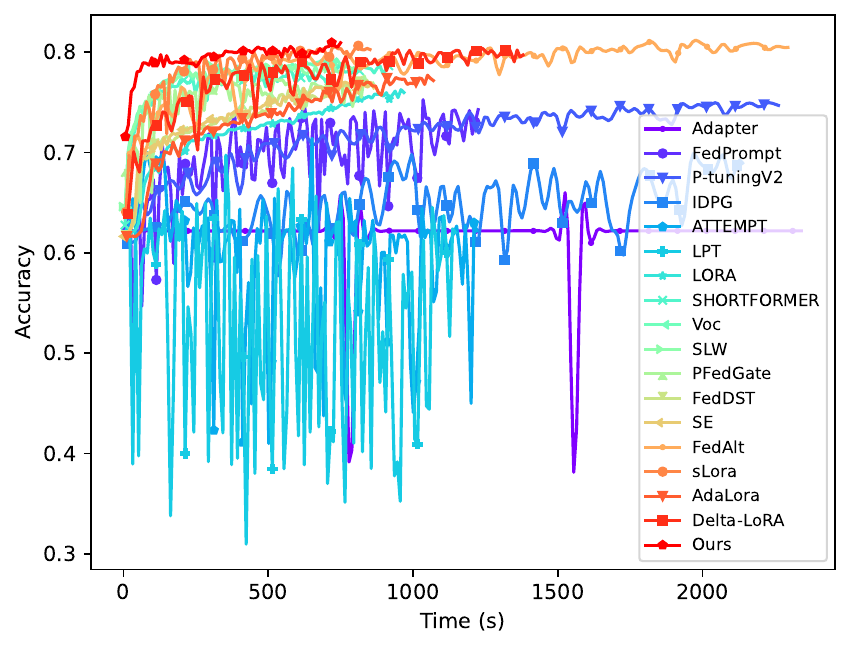}
\label{fig:boolq}
} 
\vspace*{-3mm}
\caption{The accuracy and training time with \TheName{} and diverse baseline approaches.}
\vspace*{-2mm}
\label{fig:dmrpc_rte_boolq}
\end{figure*}

\begin{figure*}[!t]
\centering
\subfigure[Acc \& MPQA]{
\includegraphics[width=0.31\textwidth]{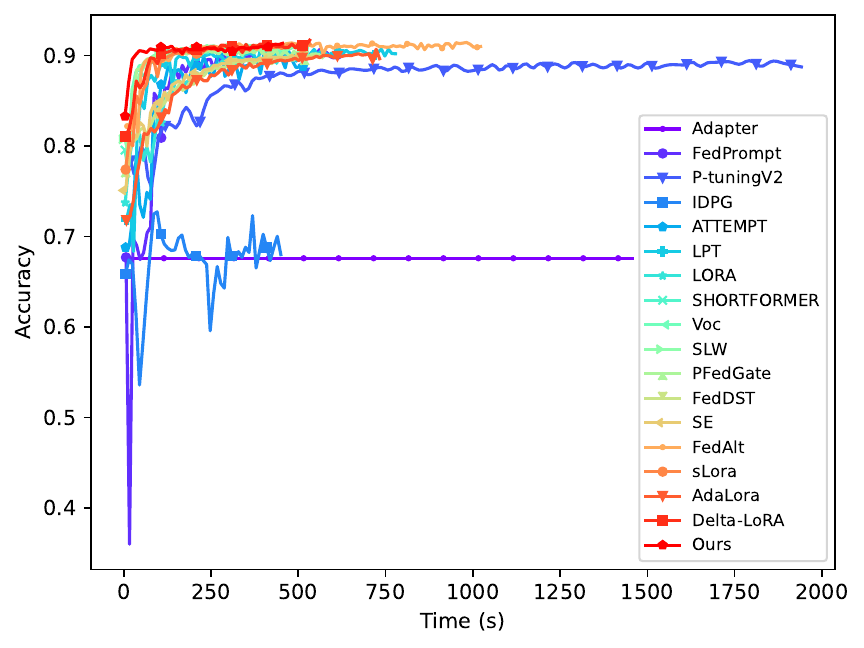}
\label{fig:mpqa}
}
\subfigure[Acc \& Subj]{
\includegraphics[width=0.31\textwidth]{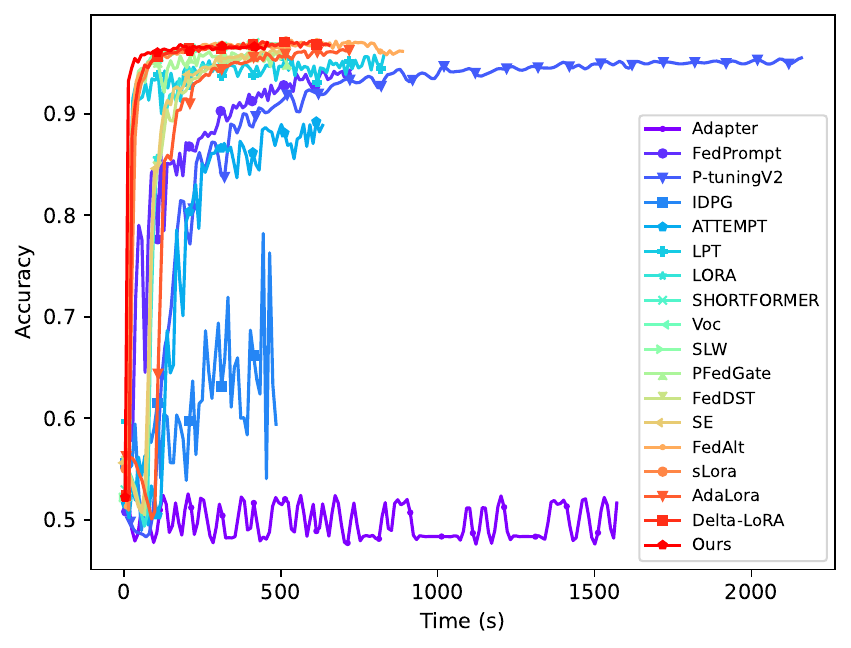}
\label{fig:subj}
}
\vspace*{-3mm}
\caption{The accuracy and training time with \TheName{} and diverse baseline approaches.}
\vspace*{-2mm}
\label{fig:mpqa_subj}
\end{figure*}

\begin{figure*}[!t]
\centering
\subfigure[Acc \& Trec]{
\includegraphics[width=0.31\textwidth]{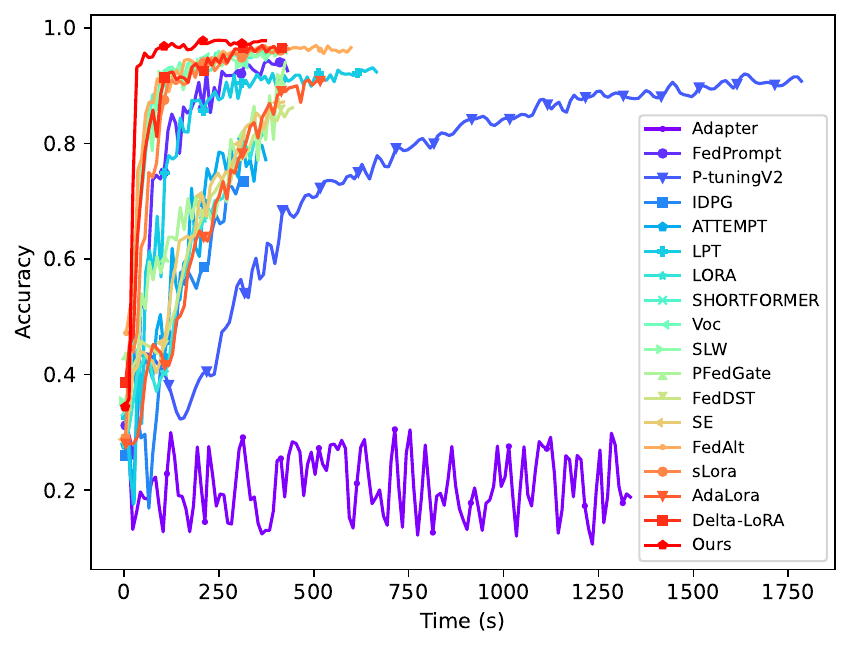}
\label{fig:trec}
}
\subfigure[Acc \& MR]{
\includegraphics[width=0.31\textwidth]{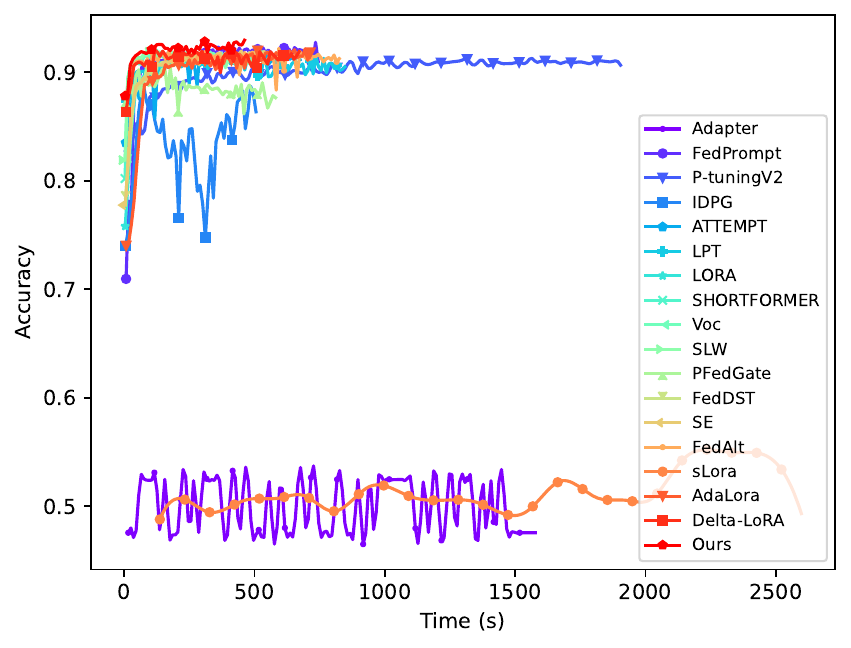}
\label{fig:mr}
} 
\vspace*{-3mm}
\caption{The accuracy and training time with \TheName{} and diverse baseline approaches.}
\vspace*{-2mm}
\label{fig:trec_mr}
\end{figure*}

Figures \ref{fig:cola_qnli_sst2} - \ref{fig:trec_mr} present the convergence process of \TheName{} and other baselines on COLA, QNLI, SST-2, MRPC, RTE, BOOLQ, MPQA and Subj datasets.
Figure \ref{fig:cola_qnli_sst2} shows the convergence results over \TheName{} and benchmarks on COLA, QNLI and SST-2 datasets, Figure \ref{fig:dmrpc_rte_boolq} presents the results on MRPC, RTE and BOOLQ datasets, Figure \ref{fig:mpqa_subj} demonstrates the evaluation results on MPQA, Subj datasets and Figure \ref{fig:trec_mr} shows the performance on Trec and MR datasets.

\subsection{Accuracy \& Time for Robustness \& Scalability}\label{subsec:robustness}
\begin{figure*}[!t]
\centering

\subfigure[Acc \& Lr]{
\includegraphics[width=0.31\textwidth]{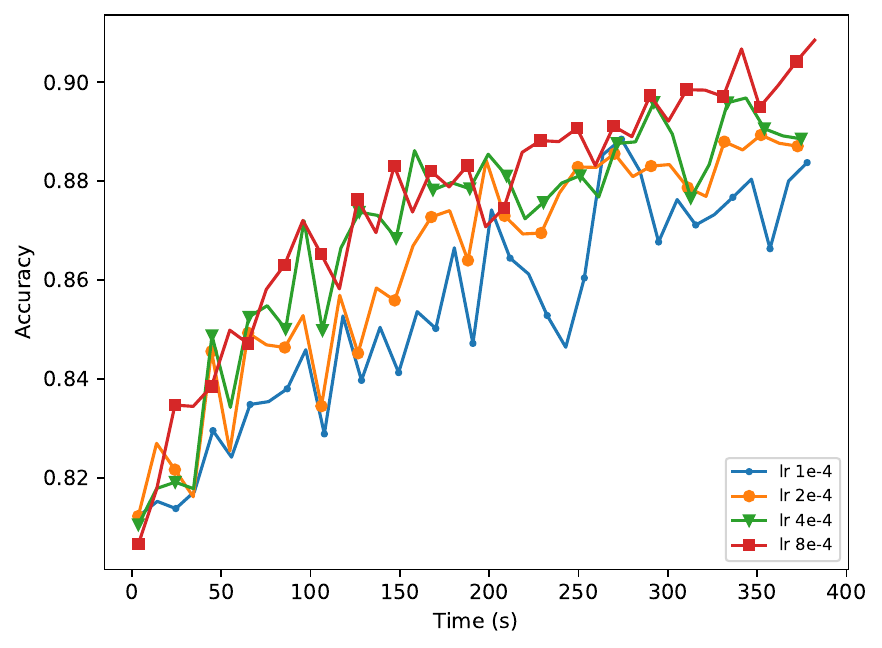}
\label{fig:lr}
}
\subfigure[Acc \& \# of devices]{
\includegraphics[width=0.31\textwidth]{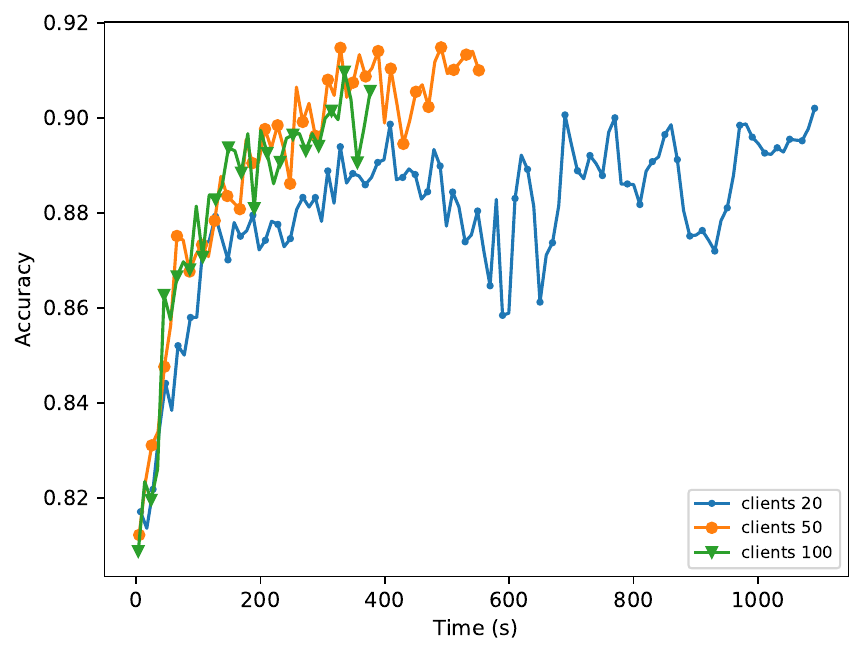}
\label{fig:clients}
}
\subfigure[Acc \& Data heterogeneity]{
\includegraphics[width=0.31\textwidth]{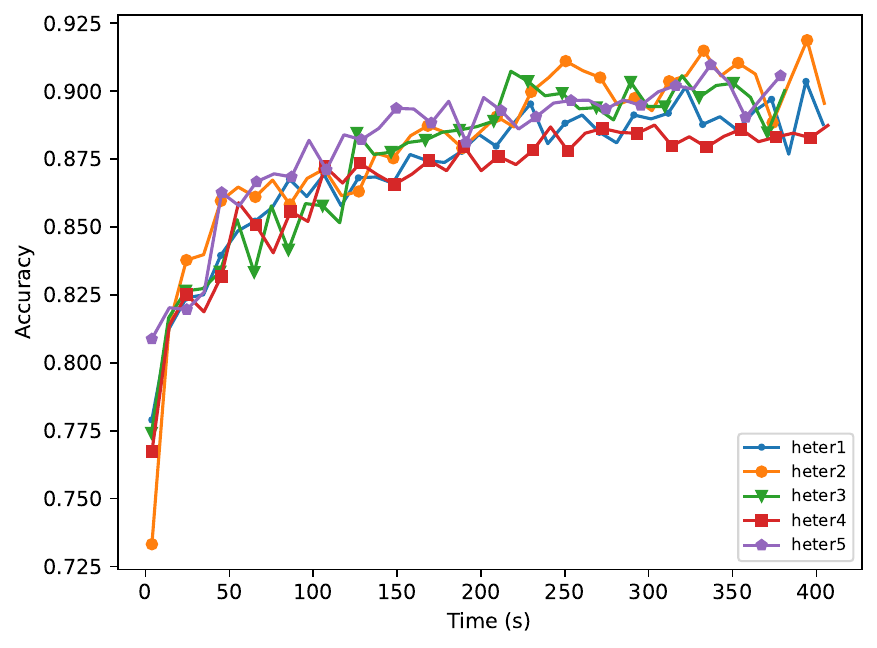}
\label{fig:data_heter}
}
\vspace*{-3mm}
\caption{The accuracy and training time with \TheName{} under different settings}
\vspace*{-2mm}
\label{fig:accuracyTime}
\end{figure*}

Figure \ref{fig:accuracyTime} presents results of robustness of \TheName{}. Figure \ref{fig:lr} shows accuracy of \TheName{} with different learning rate. Figure \ref{fig:clients} shows the performance under varying client number. In addition, Figure \ref{fig:data_heter} indicates \TheName{} is robust to different degree of heterogeneity of data.

\begin{table*}[h]
\centering
\resizebox{0.85\linewidth}{!}{
\begin{tabular}{ccccccccccc}
\toprule
\textbf{Method}    & \textbf{QNLI}      & \textbf{SST-2}       & \textbf{CoLA}         & \textbf{MPRC}       & \textbf{RTE}       & \textbf{BoolQ}       & \textbf{MPQA}         & \textbf{Subj}       & \textbf{Trec}          & \textbf{MR}   \\

\midrule

Adapter               & /  & 70  & / & 529 & / & / & /   & /  & /   & /                  \\

FedPrompt                 & 2381 & 221	& / & /	& 317	& 1121	&130	&/	&317	&95              \\
P-tuning v2                    & 1365	& 731 &894	&/	&415	&1586	&429	& 1384 & /	&992                 \\

IDPG               & /	&/	&/	&/	&/	&/	&/	&/	&\underline{73}	&/                  \\
ATTEMPT               &/	&/	&/	&/	&/	&/	&361	&/	&/	&125                \\
LPT                    & 697	& 279	&600	&/	&82	&/	&92	&246	&/	&177      \\

LORA                   & 663	&276	&230	&193	&110	&665	&210	&278	&96	&130                  \\   
SHORTFORMER                  & 193	&359	&\underline{65}	&76	&17	&66	&\underline{31}	&94	&204	&38\\ 

VOC                  & 111	&99	&71	&89	&\underline{14}	&45&32	&71	&142	&\underline{31} \\ 

SLW                   & 113	&194	&69	&60	&17	&45	&37	&71	&178	&37\\ 

PFedGate                   & \underline{85}	& /	& 110	&146	&28	&71	&33	&70	&421	&82\\ 

FedDST                   & 686	&105	&/	&/	&268	&427  	&228	&315	&/	&189\\ 

SE                   & 100	&153	&115	&/	&230	&378	&194	&283	&/	&198\\

FedALT                   & 314	&131	&288	&127	&48	&108	&46	&\underline{69}	&157	&39\\ 

sLORA                   & 170	&64	&145	&90	&45	&\underline{72}	&69	&84	&189	&/\\ 

adaLORA                   & 875	&271	&738	&/&296	&507	&265	&344	&/	&227\\ 

Delta-LoRA                   & 201	&\underline{58}	&180	&\underline{59}	&50	&123	&46	&75	&188	&66\\
\hdashline
Ours                   & \textbf{61}	&\textbf{39}	&\textbf{50}	&\textbf{28}	&\textbf{8}	&\textbf{22}	&\textbf{14}	&\textbf{25}	&\textbf{46}	&\textbf{29}\\

\bottomrule
\end{tabular}
}
\caption{The fine-tuning time (s) to achieve a target accuracy (87\% for QNLI, 93\% for SST-2, 50\% for CoLA, 83\% for MRPC, 70\% for RTE, 74\% for BoolQ, 88\% for MPQA, 95\% for Subj, 94\% for Trec, and 91.1\% for MR) with \TheName{} and diverse baseline approaches. "/" represents that fine-tuning does not achieve the target accuracy. The best results are highlighted in \textbf{bold} and the second best ones are marked with \underline{underline}. The results are obtained using RoBERTa\textsubscript{LARGE}.}
\label{tab:time_roberta}
\end{table*}

\begin{table*}[h]
\centering
\resizebox{0.85\linewidth}{!}{
\begin{tabular}{ccccccccccc}
\toprule
\textbf{Method}    & \textbf{QNLI}      & \textbf{SST-2}       & \textbf{CoLA}         & \textbf{MPRC}       & \textbf{RTE}       & \textbf{BoolQ}       & \textbf{MPQA}         & \textbf{Subj}       & \textbf{Trec}          & \textbf{MR}   \\

\midrule

batch size               & 8  & 8  & 8 & 8 & 8 & 8 & 8   & 8  & 8   & 8                  \\

non-IID degree                 & 5 & 5	& 5 & 5	& 5	& 5	5	&5	&5	&5              \\
epochs                    & 20	& 20 &100	&100	&100	&100	&100	& 100 & 100	&100                 \\

local iteration               & 2	&2	&2	&2	&2	&2	&2	&2	&2	&2                  \\
learning rate               &4e-4	&4e-4	&3e-4	&8e-4	&4e-4	&8e-4	&2e-4	&2e-4	&8e-4	&1e-4                \\
$\beta$                    & 0.6	& 0.6	&0.6	&0.6	&0.6	&0.6	&0.6	&0.6	&0.6	&0.6      \\

pace function                   & linear	&linear	&linear	&linear	&linear	&linear	&linear	&linear	&linear	&linear                  \\   
number of layers in GAL                  & 18	&18	&18	&18	&18	&18	&18	&18	&18	&18\\ 

clients num                  & 100	&100	&100	&100	&100	&100&100	&100	&100	&100 \\ 

clients num per round                   & 10	&10	&10	&10	&10	&10	&10	&10	&10	&10\\ 

$\alpha$                   & 0.8	& 0.8	& 0.8	&0.8	&0.8	&0.8	&0.8	&0.8	&0.8	&0.8\\ 

\bottomrule
\end{tabular}
}
\caption{The hyperparameters settings for each dataset}
\label{tab:exp_setup}
\end{table*}

\begin{figure*}[!t]
\centering

\subfigure[Acc \& Score function]{
\includegraphics[width=0.31\textwidth]{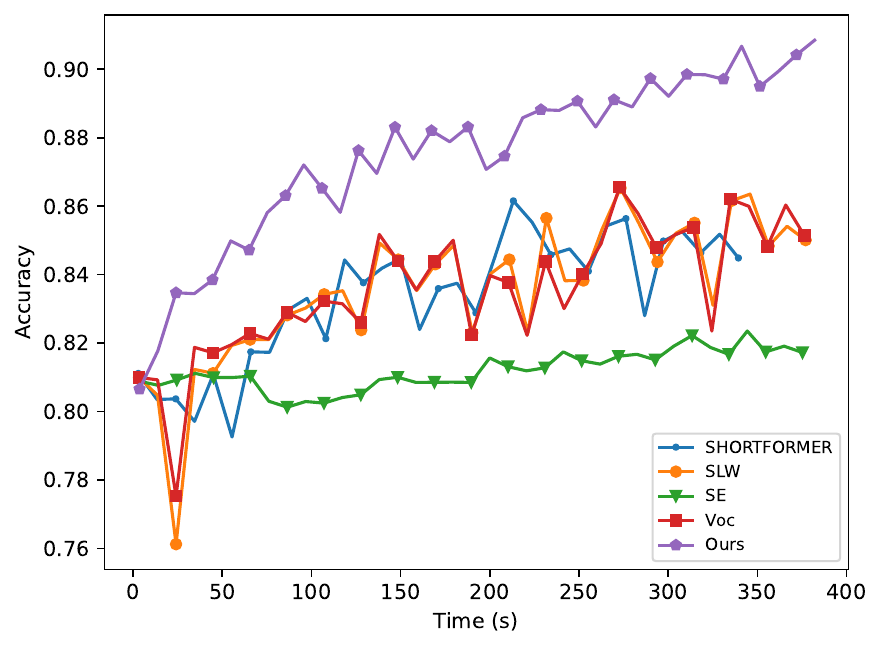}
\label{sub:cifar_resnet}
}
\subfigure[Acc \& \# of transfer layers]{
\includegraphics[width=0.31\textwidth]{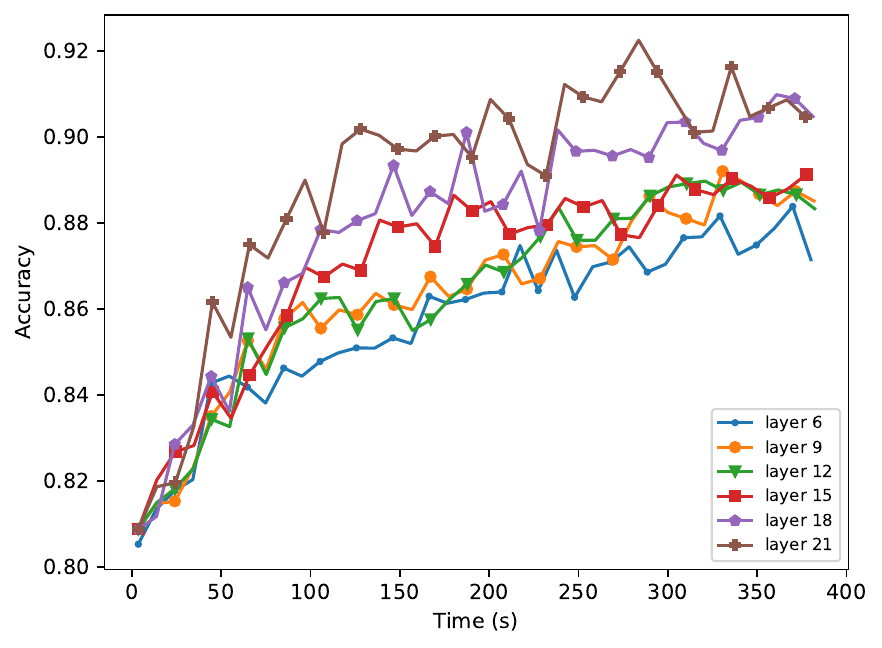}
\label{fig:layerTransfer}
}
\subfigure[Acc \& Curriculum strategy]{
\includegraphics[width=0.31\textwidth]{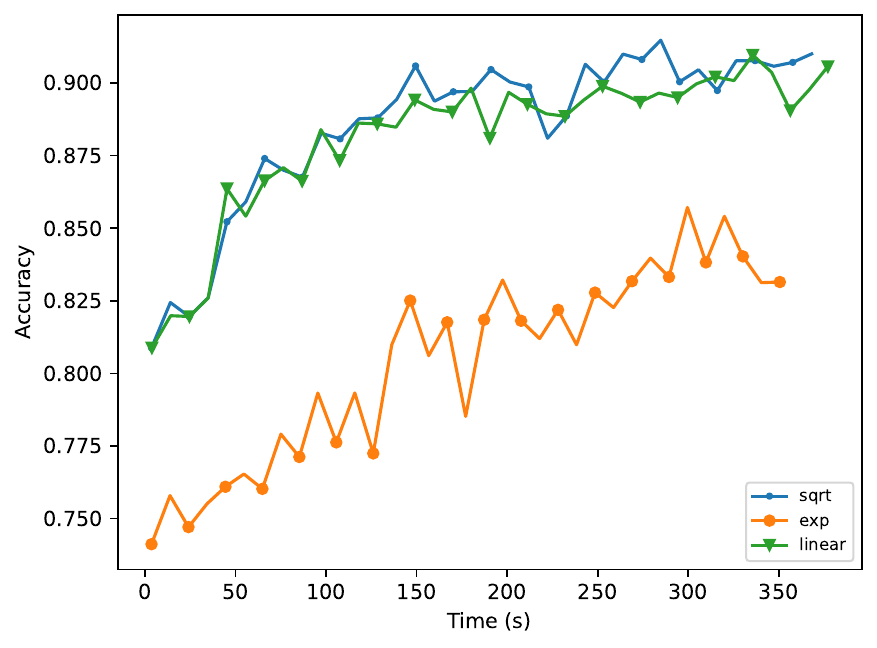}
\label{fig:pace}
}
\vspace*{-3mm}
\caption{The accuracy and training time with \TheName{} under different settings}
\vspace*{-2mm}
\label{fig:pace_score}
\end{figure*}

\begin{figure*}[!t]
\centering

\includegraphics[width=0.31\textwidth]{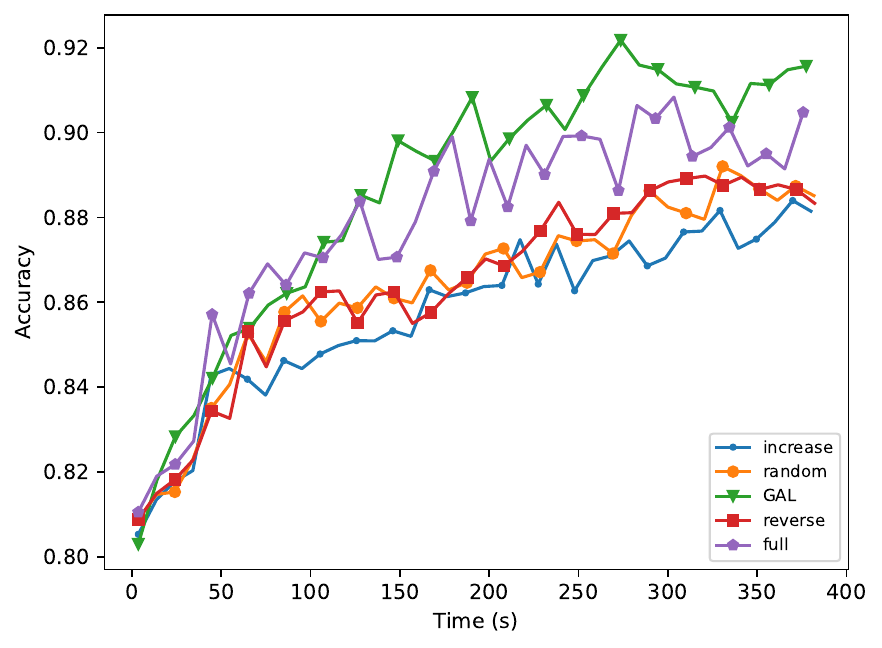}
\vspace*{-3mm}
\caption{The accuracy and training time with \TheName{} with diverse data selection strategies.}
\vspace*{-2mm}
\label{fig:dataHeter_gal}
\end{figure*}



\subsection{Impact of learning rate}\label{subsec:lr}

As shown in Figure \ref{fig:lr}. the performance of \TheName{} on MRPC dataset differs (up to 2.22\%) among 4 varying learning rates from $1e^{-4}$ to $8e^{-4}$. Thus, we take the learning rate of $\lambda=8e^{-4}$, which corresponds to the best accuracy 90.59\% in practice.

\subsection{Impact of Curriculum Strategies}\label{subsec:CL_stra}

The curriculum strategy controls the speed of exploiting difficult data samples during the fine-tuning process. We consider three strategies, i.e., linear, square (sqrt), and exponential (exp), which are defined in Formulas \ref{eq:linear}, \ref{eq:sqrt}, and \ref{eq:exp}, respectively. 
\begin{equation}
\label{eq:linear}
    \mathcal{B}^t_k = (\beta + (1 - \beta)\frac{t}{\alpha T})\frac{n_k}{\mathrm{B}},
\end{equation}
\begin{equation}
\label{eq:sqrt}
    \mathcal{B}^t_k = (\beta + (1 - \beta)\frac{t^2}{\alpha T})\frac{n_k}{\mathrm{B}},
\end{equation}
\begin{equation}
\label{eq:exp}
    \mathcal{B}^t_k = (\beta + (1 - \beta)\frac{e^{t}}{\alpha T})\frac{n_k}{\mathrm{B}},
\end{equation}
where $e$ represents the Euler's number. 
As shown in Figure \ref{fig:pace} the performance of linear (91\%) is similar to that of sqrt (91\%), which are much higher than exp (83.14\%). However, as sqrt incurs more complicated calculation compared with linear, we exploit the linear strategy in the paper. 

\subsection{Generalization of FibecFed}

To further evaluate the generalization of FibecFed on modest models, we conducted image classification task on the CIFAR-10 dataset using a seven-layer Multilayer Perceptron (MLP). As demonstrated in Table \ref{tab:accMLP}, our proposed method consistently achieves higher accuracy (from 2.09\% to 3.53\%) than FedAvg across all epochs. Furthermore, Table \ref{tab:timeMLP} illustrates that FibecFed corresponds to less time (from 0.78 to 2.12 times faster) than FedAvg. The result is aligned with the findings reported in the paper, which suggests that FibecFed can be effectively generalized across various machine learning models.

\begin{table*}[h]
\centering
\resizebox{0.5\linewidth}{!}{
\begin{tabular}{ccccc}
\toprule
\textbf{Method/Epoch number}    & 5      & 20     & 35        &  50    \\

\midrule

FedAvg               & 26.07\%   & 36.59\%  & 39.85\% & 41.13\%            \\

FibecFed        &   29.60\%  & 38.85\%	& 41.94\% & 43.74\%	              \\
\bottomrule
\end{tabular}
}
\caption{The accuracy of FibecFed and FedAvg with MLP across different epochs on cifar-10 dataset}
\label{tab:accMLP}
\end{table*}

\begin{table*}[h]
\centering
\resizebox{0.5\linewidth}{!}{
\begin{tabular}{cccc}
\toprule
\textbf{Method/Epoch number}    & 20      & 30     & 40    \\

\midrule

FedAvg               & 13.06   & 65.3  & 241.61           \\

FibecFed        &   4.18  & 25.05	& 135.43               \\
\bottomrule
\end{tabular}
}
\caption{The time (seconds) of FibecFed and FedAvg with MLP to achieve different target accuracy 40\% on cifar-10 dataset}
\label{tab:timeMLP}
\end{table*}

\subsection{Low Bound of \textit{K}}

In fact, there is no strict lower bound for $K$ (while $K$ should be bigger than 1 for federated learning). As curriculum learning is effective in centralized learning scenarios \cite{Bengio2009CL}, i.e., $K=1$, our method retains its efficacy even with the minimal device setting (1). When $K$ is increased, the performance remains high (or even higher because of more data is exploited for the training) as shown in Figure \ref{fig:clients}. In addition, we conducted additional experiments with varying numbers of devices, i.e., 2, 5, and 10, to verify the effect of small device number $K$. Table \ref{tab:accDevices} shows the performance in terms of accuracy across all tested scenarios on MRPC dataset, where the proposed method achieves an average accuracy of 90.56\% and a variance of 0.326. These results demonstrate the robustness of our approach across different $K$ values.

\begin{table*}[h]
\centering
\resizebox{0.3\linewidth}{!}{
\begin{tabular}{cc}
\toprule
Number of devices K    &  Accuracy    \\

\midrule

2               & 91.64\%  \\
5        &   89.48\%     \\
10        &   91.19\%    \\
20        &   90.20\%     \\
50        &   91.0\%     \\
100        &   90.56\%     \\
110        &   90.75\%     \\
\bottomrule
\end{tabular}
}
\caption{The accuracy under varying number of devices on the RoBERTA-Large model.}
\label{tab:accDevices}
\end{table*}

\subsection{Effect of Initial Sampling Rate}

As shown in Table \ref{tab:accTimeInitialSR}, a proper initial sample ratio can achieve optimal performance. when $B^{t}_{k}$ is small, the model can learn nothing from beginning. However, when $B^{t}_{k}$ is large, too many hard examples are exposed to model and might force model generated some bad gradients. Both cases will lead to a poor quality of aggregation on server side, hence degrading the final convergent performance. 

\begin{table*}[h]
\centering
\resizebox{0.3\linewidth}{!}{
\begin{tabular}{ccc}
\toprule
initial sample ratio    &  Accuracy  & Time  \\

\midrule

0.05      & 88.55\% &  139 \\
0.1        &   89.40\% &  141   \\
0.2        &   89.18\% &  147   \\
0.4        &   89.93\% &   162     \\
0.6        &   90.57\% &   175     \\
1.0        &   89.80\% &   205     \\
\bottomrule
\end{tabular}
}
\caption{The accuracy and training time(seconds) under varying initial sample ratios with RoBERTA-Large model on MRPC dataset.}
\label{tab:accTimeInitialSR}
\end{table*}

\section{Analysis of FibecFed}

In this section, we present the analysis of FibecFed, including the motivation, the training efficiency, fake difficulty scores, task fairness, and the significance of each component.

\subsection{Motivation}

While stringent legal regulations are carried out to protect the security and the privacy of decentralized raw data, federated learning becomes promising to enable the collaborative training process without aggregating the raw data into a centralized data center. While LLMs are too large to be directly trained with federated learning, we propose adaptive federated curriculum learning and efficient sparse parameter update with LoRA to enable the federated learning of LLMs while reducing the communication costs so as to improve the training speed (up to  98.61\% faster) and improving the accuracy (up to 45.35\%). The use case is to enable the inference process of LLMs on edge devices, while the LoRA parameters are updated. The LoRA parameters are much smaller than the full model, the training of which is feasible on edge devices. However, the full update and aggregation of all the layers within the LoRA parameters of LLMs in traditional federated learning still takes much time and the performance is inferior. Our approach can well improve the training (LoRA fine-tuning) process and the performance in the setting of federated learning of LLMs.  

\subsection{Curriculum learning \& Efficiency of the training}

Our proposed method, i.e., efficient curriculum federated learning, can significantly improve the efficiency of the training instead of reducing the efficiency. As explained in Section 5.6, we conduct a comparative analysis among four curriculum strategies:
SLW, VOC, Shortformer, SE, and that without curriculum (NULL). FibecFed corresponds to superior performance in terms of accuracy (up to 5.73\%, 9.12\% 5.84\%, 6.41\%, 7.7\% compared with Voc, SE, SLW, Shortformer, and NULL, respectively) and efficiency (up to 26.53\%, 34.26\% 68.92\%, 68.36\%, 58.57\% compared with Voc, SE, SLW, Shortformer, and NULL, respectively). The efficient curriculum federated learning can reduce the total number of batches participating in the training process so as to reduce local training time. In addition, the curriculum federated learning still carries out the training process in parallel, i.e., the training process in each client is performed in parallel. While it may take some time to calculate the score of each batch, i.e., $\mathcal{s}_i$, the calculation is carried out in parallel on each device, and takes negligible time (less than 2.98\%) compared with the training time. Once the difficulty score is determined, each training batch is sorted in ascending order and indexed starting from 0. Once the difficulty score is calculated, the training data is selected on each device based on Formula 9 for the training process of each epoch, and the selected batches of data samples are much smaller compared with those without curriculum learning at the beginning (when $j < \mathcal{B}^t_k$). The process of batch selection takes negligible time (a few microseconds) compared with the reduced training time brought by the reduced data samples. In addition, the curriculum federated learning strategy can well improve the accuracy of the trained model.

\subsection{Curriculum Learning with Fake Difficulty Score}

The Fisher Information Matrix (FIM) is calculated and stored locally, which does not need to be transferred to the server. FIM is utilized to calculate the difficult score so as to locally select the data samples based on our curriculum learning strategy on each device. When a device tricks the training by providing updates with fake lower FIM pretending to use a simpler dataset, the curriculum learning strategy can select the simple samples among the local dataset, which can reduce the training time and improve the training efficiency as well. In this work, our approaches focuses on improving the training efficiency and reducing the communication costs in federated learning of large language models based on our efficient curriculum learning method and efficient sparse parameter update, while the security issues may be addressed in future work. 

\subsection{Task Fairness}

Our efficient curriculum federated learning method does not incur unfair process of the tasks in devices. The curriculum strategy selects the data within each selected device. It does not change the device sampling mechanism, which determines which device participates in which epoch. Our efficient curriculum federated learning method guides the model to learn from simpler to more complex samples, facilitating stable and efficient convergence. In Federated Learning (FL), the inherent data heterogeneity often results in significant gradient divergence, especially when the data heterogeneity is severe. Our curriculum federated learning method can mitigate this effect, leading to more effective model aggregation and improve overall performance. 

\subsection{Communication overhead}

The absolute communication overhead is shown in the Table \ref{tab:abcom} (with RoBERTa\textsubscript{LARGE}). The communication overhead of FibecFed is higher than that of FedPrompt (up to 3.51 times), IDPG (up to 3.3 times), and ATTEMPT (up to 1.85 times). This is expected as the these three methods are prompt tuning-based methods, which corresponds to much fewer parameters to update during the training phase compared with FibecFed. However, these three methods correspond to significantly lower performance (compared with FibecFed), i.e., low convergence accuracy (from 1.25\% to 38.68\% for FedPrompt, from 6.6\% to 53.98\% for IDPG, and from 1\% to 53.94\% for ATTEMPT), as shown in Table 1. The communication overhead of FibecFed is significantly lower than the other methods (6.25 times for Adapter, 9.67 times for P-tuning V2, 1.9 times for LPT, and 25\% for LORA, SHORTFORMER, Voc, SLW, PFedGate, FedDST, SE, FedAlt, sLora, AdaLora, Delta-LoRA). This is expected as well as FibecFed only transfers the global aggregation layers instead of the parameters of all the layers in LLM. 

\begin{table*}[ht!]
\large
\centering
\resizebox{0.9\linewidth}{!}{
\begin{tabular}{cccccccccccc}
\toprule
\textbf{Method} &  \textbf{qnli} & \textbf{sst-2} &  \textbf{cola} & \textbf{mrpc} & \textbf{rte} & \textbf{boolq} & \textbf{mpqa} & \textbf{subj} & \textbf{trec} & \textbf{mr} \\
\textbf{Adapter} & 217.4 & 217.4 & 1086.9 & 1086.9 & 1086.9 & 1086.9 & 1086.9 & 1086.9 & 1086.9 & 1086.9  \\
\textbf{FedPrompt} & 6.7 & 6.7 & 33.3 & 33.3 & 33.3 & 33.3 & 33.3 & 33.3 & 33.3 & 33.3  \\
\textbf{P-tuning v2} & 320.0 & 320.0 & 1600.0 & 1600.0 & 1600.0 & 1600.0 & 1600.0 & 1600.0 & 1600.0 & 1600.0  \\
\textbf{IDPG} & 7.0 & 7.0 & 34.9 & 34.9 & 34.9 & 34.9 & 34.9 & 34.9 & 34.9 & 34.9 \\
\textbf{ATTEMPT} & 10.5 & 10.5 & 52.6 & 5.26 & 52.6 & 52.6 & 52.6 & 52.6 & 52.6 & 52.6 \\
\textbf{LPT} 87.0 & 87.0 & 434.8 & 434.8 & 434.8 & 434.8 & 434.8 & 434.8 & 434.8 & 434.8 & 434.8 \\
\textbf{LORA} 40.0 & 40.0 & 200.0 & 200.0 & 200.0 & 200.0 & 200.0 & 200.0 & 200.0 & 200.0 & 200.0 \\
\textbf{Shortformer} & 40.0 & 40.0 & 200.0 & 200.0 & 200.0 & 200.0 & 200.0 & 200.0 & 200.0 & 200.0 \\
\textbf{VOC} & 40.0 & 40.0 & 200.0 & 200.0 & 200.0 & 200.0 & 200.0 & 200.0 & 200.0 & 200.0 \\
\textbf{SLW} & 40.0 & 40.0 & 200.0 & 200.0 & 200.0 & 200.0 & 200.0 & 200.0 & 200.0 & 200.0 \\
\textbf{PFedGate} & 40.0 & 40.0 & 200.0 & 200.0 & 200.0 & 200.0 & 200.0 & 200.0 & 200.0 & 200.0 \\
\textbf{FedDST} & 40.0 & 40.0 & 200.0 & 200.0 & 200.0 & 200.0 & 200.0 & 200.0 & 200.0 & 200.0 \\
\textbf{SE} & 40.0 & 40.0 & 200.0 & 200.0 & 200.0 & 200.0 & 200.0 & 200.0 & 200.0 & 200.0 \\
\textbf{FedALT} & 40.0 & 40.0 & 200.0 & 200.0 & 200.0 & 200.0 & 200.0 & 200.0 & 200.0 & 200.0 \\
\textbf{sLORA} & 40.0 & 40.0 & 200.0 & 200.0 & 200.0 & 200.0 & 200.0 & 200.0 & 200.0 & 200.0 \\
\textbf{adaLORA} & 40.0 & 40.0 & 200.0 & 200.0 & 200.0 & 200.0 & 200.0 & 200.0 & 200.0 & 200.0 \\
\textbf{Delta-LoRA} & 40.0 & 40.0 & 200.0 & 200.0 & 200.0 & 200.0 & 200.0 & 200.0 & 200.0 & 200.0 \\
\textbf{FibecFed (ours)} & 30.0 & 30.0 & 150.0 & 150.0 & 150.0 & 150.0 & 150.0 & 150.0 & 150.0 & 150.0 \\
\bottomrule
\end{tabular}
}
\caption{Absolute communication overhead of \TheName{} and diverse baseline approaches. The time unit is second.}
\vspace{-3mm}
\label{tab:abcom}
\end{table*}

In addition, the relative communication overhead (i.e., the ratio between the absolute communication overhead and the total training time) is shown in Table \ref{tab:recom}. Similar to the absolute communication overhead, the relative communication overhead of FibecFed is higher than that of FedPrompt (from 2.3\% to 37.4\%), IDPG (from 2.4\% to 34.4\%), and ATTEMPT (from 1.9\% to 31.7\%) as well. This is expected as explained before. As the total training time of FibecFed becomes shorter, the relative communication overhead of FibecFed becomes slightly more significant than FedAlt (from 1.4\% to 20.1\%) and AdaLora (from 0.8\% to 5.7\%). In addition, the relative communication overhead of FibecFed becomes similar to that of sLora (from 7.8\% smaller to 24.8\% bigger) and Delta-LoRA (from 7.5\% smaller to 5.5\% bigger).  The relative communication overhead of FibecFed is still smaller than the rest approaches, i.e., Adapter (up to 43.9\%), P-tuning V2 (up to 55.6\%), LPT (up to 25.1\%), LORA (up to 14.5\%), SHORTFORMER (up to 13.8\%), Voc (up to 8.0\%), SLW (up to 8.8\%), PFedGate (up to 7.0\%), FedDST (up to 7.1\%), SE (8.1\%). Please note that FibecFed corresponds to higher convergence accuracy (as shown in Table 1 in the manuscript) and shorter training time (as shown in Table 2 in the manuscript).

\begin{table*}[ht!]
\large
\centering
\resizebox{0.9\linewidth}{!}{
\begin{tabular}{cccccccccccc}
\toprule
\textbf{Method} &  \textbf{qnli} & \textbf{sst-2} & \textbf{cola} & \textbf{mrpc} & \textbf{rte} & \textbf{boolq}  &  \textbf{mpqa} &  \textbf{subj} &  \textbf{trec}  &  \textbf{mr} \\
\textbf{Adapter} & 0.091 & 0.161 & 0.674 & 0.837 & 0.833 & 0.464 & 0.746 & 0.692 & 0.815 & 0.689 \\
\textbf{FedPrompt} & 0.003 & 0.005 & 0.046 & 0.078 & 0.097	& 0.027 & 0.054 & 0.046 & 0.077 & 0.045 \\
\textbf{P-tuning v2} & 0.127 & 0.260 & 0.858 & 0.899 & 0.883 & 0.707 & 0.824 & 0.741 & 0.897 & 0.840 \\
\textbf{IDPG} & 0.003 & 0.005 & 0.067 & 0.107 & 0.096 & 0.016 & 0.078 & 0.072 & 0.111 & 0.069 \\
\textbf{ATTEMPT} & 0.004 & 0.010 & 0.088 & 0.135 & 0.140 & 0.043 & 0.100 & 0.083 & 0.141 & 0.078 \\
\textbf{LPT} & 0.045 & 0.098 & 0.546 & 0.664 & 0.630 & 0.379 & 0.558 & 0.525 & 0.653 & 0.513 \\
\textbf{LORA} & 0.021 & 0.042 & 0.416 & 0.570 & 0.539 & 0.207 & 0.456 & 0.454 & 0.535 & 0.441 \\
\textbf{Shortformer} & 0.048 & 0.072 & 0.370 & 0.589 & 0.478 & 0.218 & 0.429 & 0.428 & 0.531 & 0.409 \\
\textbf{VOC} & 0.023 & 0.037 & 0.351 & 0.531 & 0.441 & 0.254 & 0.383 & 0.398 & 0.471 & 0.373 \\
\textbf{SLW} & 0.023 & 0.036 & 0.366 & 0.530 & 0.482 & 0.252 & 0.382 & 0.397 & 0.475 & 0.377 \\
\textbf{PFedGate} & 0.019 & 0.032 & 0.336 & 0.521 & 0.441 & 0.232 & 0.351 & 0.378 & 0.471 & 0.344 \\
\textbf{FedDST} & 0.020 & 0.034 & 0.322 & 0.523 & 0.448 & 0.238 & 0.353 & 0.384 & 0.451 & 0.350 \\
\textbf{SE} & 0.021 & 0.032 & 0.338 & 0.533 & 0.454 & 0.231 & 0.375 & 0.385 & 0.476 & 0.370 \\
\textbf{FedALT} & 0.007 & 0.015 & 0.184 & 0.251 & 0.262 & 0.087 & 0.196 & 0.225 & 0.334 & 0.242 \\
\textbf{sLORA} & 0.022 & 0.034 & 0.354 & 0.530 & 0.465 & 0.235 & 0.375 & 0.372 & 0.462 & 0.077 \\
\textbf{adaLORA} & 0.011 & 0.021 & 0.263 & 0.435 & 0.364 & 0.187 & 0.273 & 0.278 & 0.379 & 0.280 \\
\textbf{Delta-LoRA} & 0.019 & 0.037 & 0.339 & 0.527 & 0.453 & 0.145 & 0.375 & 0.306 & 0.469 & 0.298 \\
\textbf{FibecFed (ours)} & 0.019 & 0.029 & 0.302 & 0.452 & 0.394 & 0.200 & 0.330 & 0.328 & 0.403 & 0.325 \\
\bottomrule
\end{tabular}
}
\caption{Relative communication overhead of \TheName{} and diverse baseline approaches. The time unit is second.}
\vspace{-3mm}
\label{tab:recom}
\end{table*}

\subsection{Significance of Each Component}

Our approach aims to improve the efficiency of LLM federated learning in two perspective: communication cost and local training efficiency, each of which plays a significant role in accelerating the training process of FL. In contrast to heuristic metrics, we utilize the FIM to assess sample difficulty, achieving an accurate estimation of difficulty score. This newly proposed metric enhances Curriculum Learning by facilitating a more stable and faster convergence training, reducing the number of epochs required for local training. Consequently, this approach reduces local training time and improves overall efficiency, offering a significant enhancement over traditional difficulty assessment methods. In addition, A novel noise-sensitive Layer Selection is proposed to identify the critical layers of the model, thereby reducing communication overhead without compromising performance. Furthermore, A novel parameter selection strategy focuses on identifying key neurons within the model for local update. By determining which parameters are most influential, this approach enhances local training efficiency, optimizes computational resources and speeds up the learning process on individual devices. 

In Section 5.6 (Ablation Study), we demonstrate the ablation study in terms of the curriculum data selection method, the important layer selection method, and the local update parameter selection method. The experimentation reveals that the efficient curriculum federated learning corresponds to superior performance in terms of both accuracy (up 7.7\%) and efficiency (up to 68.92\%). In addition, the important layer selection method can significantly improve the efficiency (up to 23.1\%) with slight improvement of accuracy (up to 3.42\%) and the local update parameter selection method can further improve the efficiency up to 11.8\% with slightly higher efficiency (up to 2.48\%).

In our approach, efficient curriculum federated 
learning enables the training process begins with simple data samples and then gradually increase the difficulty, which can improve both the efficiency and the performance. The efficient sparse parameter update method is composed of global layer selection and local update parameter selection. The global layer selection method reduces communication costs by only transferring the layers of importance scores between devices and the server, without performance degradation. The local update parameter selection can improve the local training efficiency by only updating the important parameters. Our approach yields excellent performance (up to 45.35\% in terms of accuracy) and superb fine-tuning speed (up to 98.61\% faster).

\subsection{Combination with Model Compression}

Our approach is orthogonal with model compression methods. Various model compression methods exist, e.g., pruning and quantization. Our approach can be combined with these methods to achieve higher training speed or higher accuracy. However, the pruning and quantization methods may degrade the performance (accuracy) of LLMs. In addition, the training process based on the model compression methods may require additional low-level training operators in the forward propagation and back propagation, which incurs extra complexity. Thus, the combination of our approach with the model compression methods can be addressed in our future work.

\subsection{LLM Training on Devices}

In real-world scenarios, numerous governments and organizations have established regulations and laws to protect data privacy, making access to data on edge devices increasingly challenging. Moreover, pre-trained models are often not well-suited for domain-specific tasks due to lack of fine-tuning. By implementing our proposed approach, the LLMs can be fined based on the distributed raw data stored on diverse edge devices without aggregating the raw data into a central server or a central data center. This approach not only adheres to privacy and regulatory standards but also ensures that LLMs can be fine-tuned to specific tasks with distributed end user data distributed in edge devices. A classical example is highlighted in \cite{zhao2024llmbased}, where, by leveraging the behavioral data of end users, the server obtains a collaboratively fine-tuned LLM-based recommendation system. Furthermore, with the advancement of LLM technology, additional applications could emerge in highly confidential environments such as hospitals and banks. Our proposed approach provides a solution to develop good task-specific model while ensuring both the integrity and confidentiality of the sensitive raw data of end users on diverse edge devices. All the explanation will be added in our final version. 

\subsection{Data Heterogeneity \& Layer Importance}

The inherent heterogeneity of samples on each device contributes to the difference of the importance scores of each layer. In the settings of federated learning, the samples on each device are generally heterogeneous, i.e., non non-Independent and Identically Distributed (non-IID). In order to select the important layers as the global aggregation layers, we aggregate the importance scores on each device based on Formula \ref{eq:layerScoreGlobal} so as to calculate the global importance scores for each layer. This global aggregation can balance the diverse importance of each layer on each device. Afterward, we select the proper global aggregation layers with a lossless method on each device. This selection can reduce the data to communicate in each epoch so as to improve the efficiency without degrading performance (accuracy).

\subsection{Perturbed Parameters \& Increased Sensitivity}

While the layer selection is conducted on each device, the importance scores of each layer are aggregated, which can balance the importance of layers among different devices. The data heterogeneity may have impact on the global aggregation layer selection within the lossless method, i.e., the number of selected aggregation layers are different across devices. 

When more parameters are perturbed locally, the lossless method may result in more global aggregation layers so as to achieve excellent performance (accuracy). When the sensitivity to noise corresponding to specific layers is increased, the impact can be taken into consideration while calculating the importance scores. Afterward, the impact is reflect within the calculation of Formula 15 to generate the global importance score of each layers. Thus, the increase sensitivity to noise may impact the importance of layers. However, the fact of more parameters are perturbed can be captured by our lossless method, which determines a proper number of layers to be selected as the global aggregation layers so as to reduce the number of layers to transfer while ensuring the performance (accuracy) of LLM in federated learning. More parameters are perturbed locally across different clients when their increased sensitivity to noise has impacts on more layers of LLMs. Our proposed global aggregation layer selection with the lossless method can well reduce the communication costs so as to improve the efficiency (training speed) while ensuring the performance (accuracy).

\subsection{Implementation of \TheName{}}

\TheName{} is composed of two methods, i.e., adaptive federated curriculum learning and efficient sparse parameter update. While these two methods correspond to in-depth novel technical contributions based on Fisher Information, the implementation of \TheName{} is straightforward. While Fisher Information is exploited, we calculate the square of the elements in the diagonal of the first-order derivative matrix to reduce the complexity of the Fisher Information computation as explained in Section \ref{subsec:cl}. 

In practice, the Fisher Information is implemented to calculate the importance scores of the data and layers in the initialization phase. In addition, the local update parameter selection is carried out in the initialization phase as well. Within the training phase, the curriculum data selection strategy is carried out. The execution can be easily carried out with the parameters, e.g., $\alpha$, $\beta$, as shown in Table \ref{tab:exp_setup}. 

\subsection{intuition for Fisher Information}

Within the training phase of federated learning, it is pivotal to exploit the informative data and to update the critical parts of the LLM so as to achieve efficient training process. However, the existing methods to evaluate the training data or the LLM is static, which corresponds to inaccurate estimation and low performance. As a measure of the local curvature, Fisher Information Matrix (FIM) defines the Riemannian metric of the parameter space, which can indicate the difficulty of data samples and the importance of each component of the LLM within the training phase. Fisher information is defined as how sensitive the model is to changes in $\theta$ at a particular $\theta$ \cite{ly2017tutorial}, where $\theta$ represents the value to infer and corresponds to the ground truth output of the LLM in FibecFed. When the Fisher Information is not significant in respect of a certain training sample within the training process of an LLM, i.e., small score defined in Formula \ref{eq:score_batch}, the corresponding training data contains relatively few information for the training process and the data is considered easy for the LLM. Then, the model can quickly learn the knowledge within the easy data according to the starting small strategy of the curriculum learning \cite{Bengio2009CL}. In FibecFed, we exploit Formula \ref{eq:cl} to carry out the Fisher Information-based curriculum learning, which can choose easy samples at the beginning to improve the training efficiency while achieving high accuracy. Besides the training data, Fisher Information can indicate the importance of the neuron in LLM within the training phase. When the Fisher Information corresponding to a neuron is significant, we consider it as an important part of the LLM that needs further adjustment (training). Otherwise, the corresponding neurons do not need update as they are not sensitive within the training phase (these neurons may stay unchanged even when they are considered as local update parameters). Thus, we take the neurons of high importance scores based on the Fisher Information (based on Formula \ref{eq:neuron_sum}) as the local update parameters. In this way, Fisher Information guides FibecFed to choose the simple data to begin with and to choose the sensitive neurons in the LLM to update within the training phase so as to achieve efficient and effective federated learning.

\subsection{Privacy in \TheName{}}


Within the initialization of \TheName{}, only the global aggregation layer selection incurs exchanging the importance scores of each layer in the LLM between the server and devices, while the other two modules, i.e., Fisher information-based curriculum learning and local update parameter selection, does not bring any extra information transfer between the server and devices. The importance scores of layers are insensitive data with few information of the raw training data on each device. The insensitive data resembles the number of samples to be transferred in traditional federated learning approaches, e.g., FedAvg, which still complies with privacy regulations. To the best of our knowledge, there is no attack methods based on the importance scores of each layer in the LLM.

\TheName{} can be combined with other security or privacy methods, e.g., encryption, differential privacy, to further protect the data security and privacy in federated learning. 

\section{Potential Risks}

We propose an efficient LLM federated learning approach to enable collaborative LLM training with distributed raw data without aggregation, which can protect the privacy of the raw data. Our approach can be combined with other defense approaches or privacy protection methods to further enhance the security or the privacy of federated learning when there are curious or malicious attacks.

\end{document}